\documentclass[sigconf]{acmart}

\AtBeginDocument{%
  }

\setcopyright{none} 
\acmConference[KDD '26] % 1. 会议简称
{The 32nd ACM SIGKDD Conference on Knowledge Discovery and Data Mining}{August 2026}{Jeju, Korea}

\usepackage{booktabs}
\usepackage{algorithm}
\usepackage{algpseudocode}
\usepackage{amsmath}

\usepackage{threeparttable}
\usepackage{siunitx}
\usepackage{enumitem}  
\usepackage{multirow}

\begin{document}
\settopmatter{printacmref=false} % 隐藏 ACM 参考格式块
\renewcommand\footnotetextcopyrightpermission[1]{} % 隐藏版权声明

\title{DSPR: Dual-Stream Physics-Residual Networks for Trustworthy Industrial Time Series Forecasting}

%% --- 第一作者 (两个单位) ---
\author{Yeran Zhang}
\affiliation{%
  \institution{School of Computer Science and Engineering, University of Electronic Science and Technology of China}
  \city{Chengdu}
  \state{Sichuan}
  \country{China}
}
\affiliation{%
  \institution{Research Center, East Hope Group Co., Ltd}
  \city{Shanghai}
  \country{China}
}
\email{yeranzhang36@gmail.com}

%% --- 第二作者 ---
\author{Pengwei Yang}
\affiliation{%
  \institution{School of Computer Science and Engineering, University of Electronic Science and Technology of China}
  \city{Chengdu}
  \state{Sichuan}
  \country{China}
}
\email{pengwei.yang@std.uestc.edu.cn}

%% --- 第三作者 ---
\author{Guoqing Wang}
\affiliation{%
  \institution{School of Computer Science and Engineering, University of Electronic Science and Technology of China}
  \city{Chengdu}
  \state{Sichuan}
  \country{China}
}
\email{gqwang0420@uestc.edu.cn}

%% --- 通讯作者 ---
\author{Tianyu Li}
\authornote{Corresponding author} 
\affiliation{%
  \institution{School of Computer Science and Engineering, University of Electronic Science and Technology of China}
  \city{Chengdu}
  \state{Sichuan}
  \country{China}
}
\email{tylisky@uestc.edu.cn}


\begin{abstract}

Accurate forecasting of industrial time series requires balancing predictive accuracy with physical plausibility under non-stationary operating conditions. Existing data-driven models often achieve strong statistical performance but struggle to respect regime- dependent interaction structures and transport delays inherent in real-world systems. To address this challenge, we propose DSPR (Dual-Stream Physics–Residual Networks), a forecasting framework that explicitly decouples stable temporal patterns from regime-dependent residual dynamics. The first stream models the statistical temporal evolution of individual variables. The second stream focuses on residual dynamics through two key mechanisms: an Adaptive Window module that estimates flow-dependent transport delays, and a Physics-Guided Dynamic Graph that incorporates physical priors to learn time-varying interaction structures while suppressing spurious correlations. Experiments on four industrial benchmarks spanning heterogeneous regimes demonstrate that DSPR consistently improves forecasting accuracy and robustness under regime shifts while maintaining strong physical plausibility. It achieves state-of-the-art predictive performance, with Mean Conservation Accuracy exceeding 99\% and Total Variation Ratio reaching up to 97.2\%. Beyond forecasting, the learned interaction structures and adaptive lags provide interpretable insights that are consistent with known domain mechanisms, such as flow-dependent transport delays and wind-to-power scaling behaviors. These results suggest that architectural decoupling with physics-consistent inductive biases offers an effective path toward trustworthy industrial time-series forecasting. Furthermore, DSPR's demonstrated robust performance in long-term industrial deployment bridges the gap between advanced forecasting models and trustworthy autonomous control systems.

\end{abstract}


\maketitle

% ================= SECTION 1: INTRODUCTION =================

\section{Introduction}
\label{sec:intro}

In the era of AI for Science, forecasting complex industrial systems confronts a fundamental tension. First-principles models such as differential equations offer interpretability and strict adherence to conservation laws but often fail to capture stochastic nuances of real-world data due to simplified assumptions~\cite{QIN2003733,camacho2007}. 
Conversely, data-driven Deep Learning models, particularly Transformers~\cite{zhou2021,wu2023}, achieve remarkable predictive accuracy yet remain physically blind black boxes. 
In safety-critical settings including emission control and power dispatch, this opacity poses severe risk: a model that minimizes Mean Squared Error while violating mass balance or thermodynamic causality is fundamentally untrustworthy.

\begin{figure}[h]
    \centering
    \includegraphics[width=\linewidth]{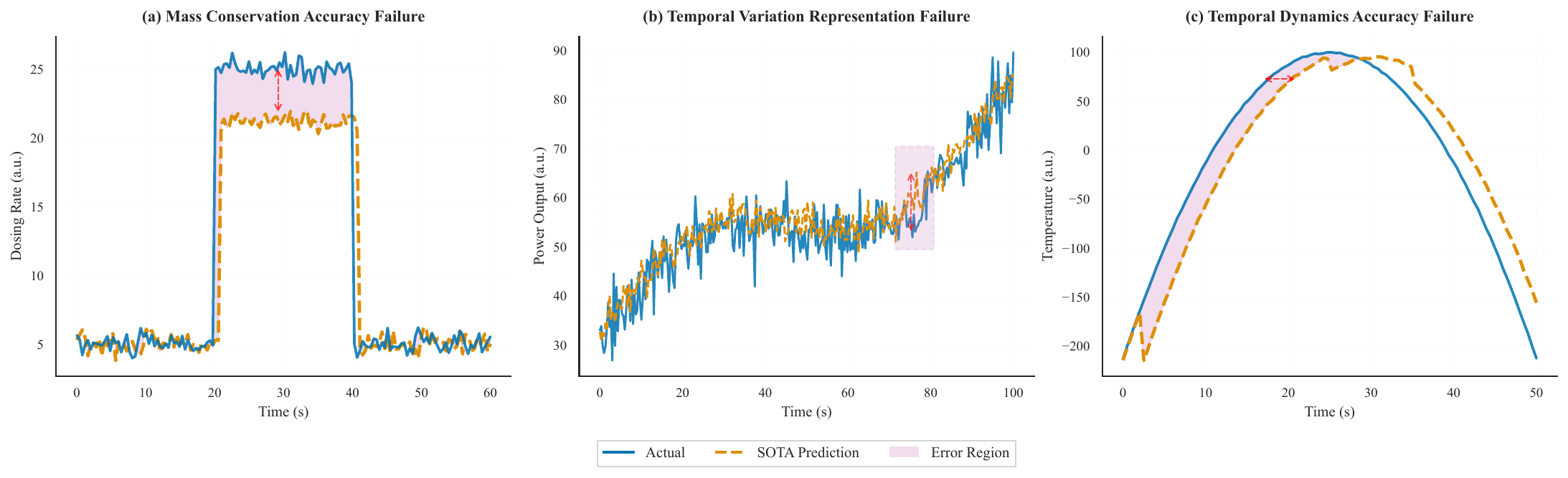}
    \caption{\textbf{Fidelity collapse in state-of-the-art industrial forecasting.} 
    (a) \textbf{MCA Failure}: SOTA models fail to capture abrupt step responses, violating local mass conservation (shaded areas). 
    (b) \textbf{TVR Failure}: Statistical averaging over-smooths high-frequency transients, suppressing critical physical dynamics. 
    (c) \textbf{TDA Failure}: Predictive lags at regime transitions yield incorrect causal directions (directional mismatch).}
    \label{fig:fidelity_failure_cases}
\end{figure}

The challenge is exacerbated by regime-dependent dynamics inherent in industrial processes~\cite{skaf2014,yang2022}. 
Unlike stationary time series, physical systems exhibit time-varying characteristics driven by operating conditions. 
Variable transport delays in fluid-driven systems cause lags between actuation and response to fluctuate with flow velocity, rendering static assumptions invalid. 
Non-stationary couplings shift dominant dependencies dynamically, where heat transfer limitations may dominate at high loads while reaction kinetics govern low-load states.
Standard DL models~\cite{Han2021GTS}, lacking structural priors, struggle to distinguish valid physical shifts from sensor noise. 
As illustrated in Fig.~\ref{fig:fidelity_failure_cases}, SOTA forecasters suffer from notable fidelity collapse across multiple dimensions: failing to capture abrupt step responses (violating mass conservation), over-smoothing high-frequency transients (suppressing critical dynamics), and introducing predictive lags at regime transitions (yielding incorrect causal directions). While achieving low statistical error (MAE/RMSE), these models sacrifice physical plausibility for numerical precision.

To bridge this accuracy-fidelity dilemma, we propose DSPR (Dual-Stream Physics-Residual Networks), a framework that fundamentally shifts physics integration from passive soft constraints in Physics-Informed Neural Networks to active architectural inductive biases. 
DSPR decomposes dynamics into a \textbf{Trend Stream} that absorbs high-energy inertial patterns, thereby isolating subtle, physics-governed transients into the \textbf{Residual Stream} for focused constraint learning.
Crucially, we embed domain knowledge directly into network structure: an Adaptive Window module explicitly learns flow-dependent transport delays, while a Dynamic Graph module disentangles causal topology from spurious correlations using physical priors.

We validate DSPR on four diverse datasets spanning chemical kinetics in SCR, thermodynamics in Kiln, process control in TEP, and energy meteorology in SDWPF. Our contributions are summarized as follows:

\begin{itemize}[leftmargin=1.0em, labelsep=0.5em]
    \item \textbf{Mechanism-aligned surrogate for non-stationary physical systems.}
    We propose DSPR, a dual-stream architecture that explicitly decomposes industrial dynamics into (i) stable inertial trends and (ii) regime-dependent physical residuals. By embedding a learnable transport-delay operator (Adaptive Window) and a prior-guided dynamic interaction graph into the model structure, DSPR captures time-varying lags and couplings that standard black-box forecasters typically confound with noise.

    \item \textbf{Extracting scientific quantities from sensors for mechanism analysis.}
    DSPR exposes interpretable intermediate representations, including learned delay profiles and dynamic coupling graphs, which serve as \emph{measurable scientific quantities}. These quantities recover meaningful domain mechanisms from noisy multivariate measurements, including flow-dependent reaction lags in SCR and the wind-to-power conversion pathway consistent with aerodynamic scaling in SDWPF, supporting mechanism-level analysis beyond predictive accuracy.

    \item \textbf{Resolving the accuracy--fidelity dilemma with trustworthy downstream impact.}
    We introduce a unified evaluation of predictive precision and physical fidelity using conservation and dynamics-aware criteria (MCA/TVR/TDA)~\cite{Beucler_2021, RUDIN1992259, Hashem1992}, revealing the fidelity collapse of purely data-driven baselines. Across diverse physical regimes, DSPR achieves a stronger accuracy--fidelity balance, and its mechanism-consistent predictions enable deployment in a production-grade control workflow on a 5,000 t/d cement line with sustained safe operation and measurable resource savings \textbf{(see Appendix~\ref{app:deployment} for detailed deployment analysis and economic impact)}.
\end{itemize}

This work establishes that prior-guided architectural adaptation, rather than black-box scaling, constitutes the key to trustworthy scientific machine learning in complex industrial environments.

% ================= SECTION 2: RELATED WORK =================
\section{Related Work}
\label{sec:related_work}

\noindent\textbf{Time Series Forecasting.}
Modern forecasting has evolved from classical methods (ARIMA, VAR)~\cite{rahman2017modeling, Chris1980} to deep Transformer architectures. While earlier models prioritized efficiency and frequency decomposition~\cite{zhou2021,wu2021,zhou2022fedformer}, recent SOTA approaches—such as PatchTST~\cite{Yuqi2023}, iTransformer~\cite{liu2023}, TimeMixer~\cite{wang2024}, and TimesNet~\cite{wu2023}—focus on scalability through patching, inverted attention, and multi-scale modeling. Despite their statistical precision, these data-driven methods lack intrinsic physical constraints, often violating conservation laws and causal monotonicity in scientific applications~\cite{Kong2025,LAWRENCE2024}.

\noindent\textbf{Graph Neural Networks for Spatiotemporal Learning.}
GNNs capture spatiotemporal dependencies by integrating graph convolutions with recurrent units~\cite{yu2018,li2018dcrnn_traffic} or learning adaptive topologies~\cite{wu2020}. Recent spectral advances, such as \textbf{MSGNet}~\cite{cai2023msgnet} and \textbf{TimeFilter}~\cite{hu2025timefilter}, further exploit frequency-domain filters to efficiently model multi-scale correlations. However, these methods typically rely on assumptions of structural stability, limiting their efficacy in industrial settings where non-stationary physics drive dramatic shifts in causal structures across operating regimes~\cite{yang2022}.

\noindent\textbf{Physics-Informed Scientific Machine Learning.}
Physics- Informed Neural Networks (PINNs)~\cite{raissi2019physics,Karniadakis2021} incorporate governing equations as soft loss constraints, with extensions including conservative PINNs~\cite{cai2021physics} and neural operators~\cite{lu2021learning}. While effective for simulation, PINNs require explicit equation formulation often unavailable for complex catalytic reactions and prove fragile under noisy industrial measurements. Alternative approaches—hybrid models~\cite{willard2022}, sparse identification~\cite{BRUNTON2016710}, and graph-based reaction networks~\cite{li2018dcrnn_traffic,wu2020}—face limitations including validation primarily on synthetic data and assumptions of stable dynamics. Our work differs by embedding physical knowledge as architectural inductive biases (adaptive windows for transport delays, physics-guided graphs for reaction directionality) rather than loss penalties, enabling regime adaptation and structure discovery validated on real industrial data.

% ================= SECTION 3: METHODOLOGY =================
\section{Methodology}
\label{sec:methodology}

\begin{figure*}
  \centering
  \includegraphics[width=0.95\textwidth]{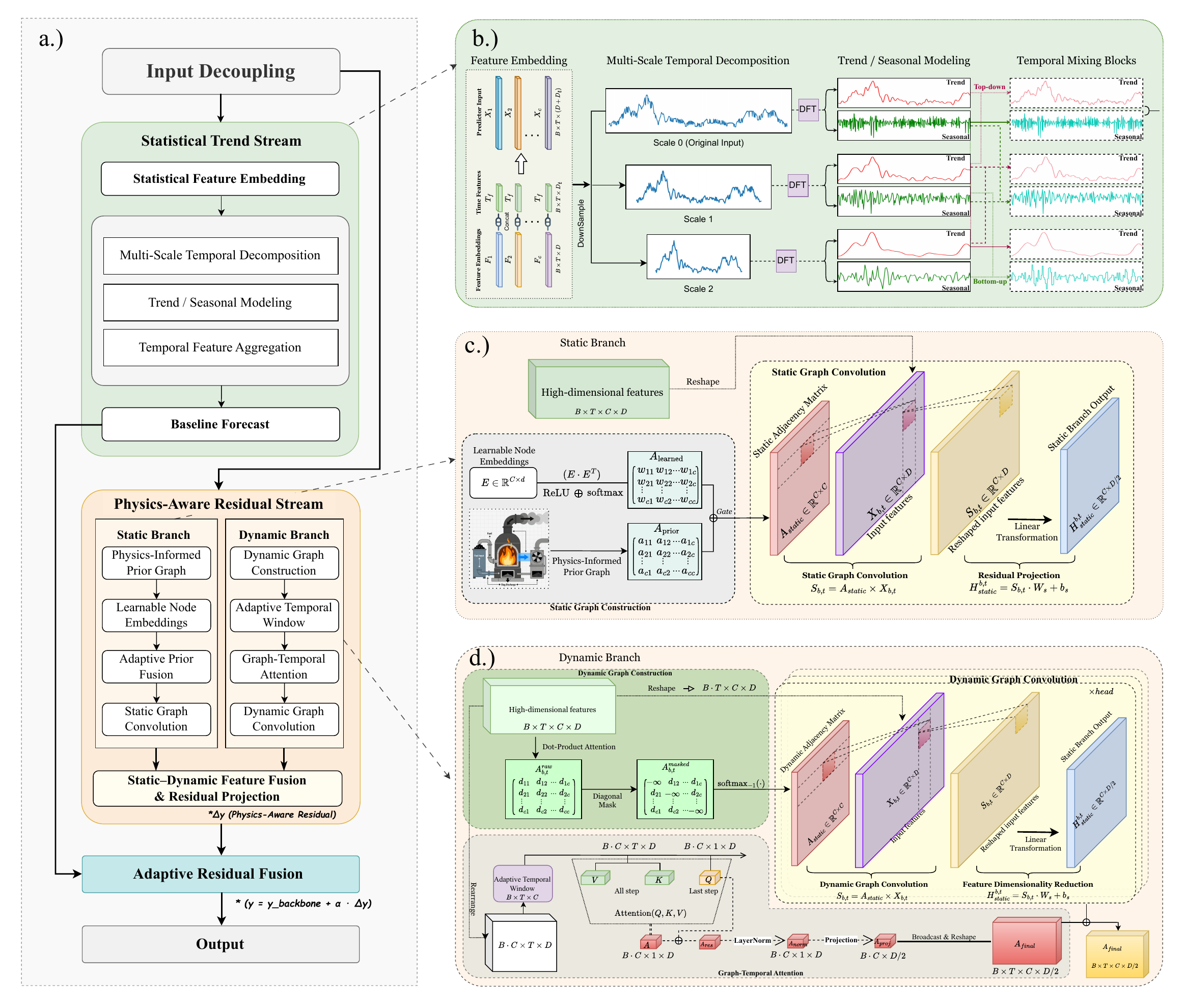}
  \caption{Overview of the proposed DSPR framework. 
  \textbf{a) Overall Architecture:} Decouples dynamics into statistical patterns and physical residuals. 
  \textbf{b) Statistical Base Stream:} Extracts dominant temporal patterns via multi-scale decomposition. 
  \textbf{c) Static Branch:} Models invariant spatial dependencies using physical priors. 
  \textbf{d) Dynamic Branch:} Captures transient fluctuations via Adaptive Windows and dynamic graphs.}
  \label{fig:architecture}
\end{figure*}

We propose the \textbf{Dual-Stream Physics-Residual Framework (DSPR)}, which forecasts non-stationary industrial dynamics by decomposing system evolution into \textbf{dominant statistical patterns} and regime-dependent local deviations. The overall architecture, comprising a \textit{Statistical Stream} and a \textit{Physics-Aware Residual Stream}, is presented in Fig. ~\ref{fig:architecture}a, with the forward propagation summarized in Algorithm~\ref{alg:forward_pass}.

\subsection{Problem Formulation}
Let $\mathcal{X} = \{ \mathbf{X}_{t-L+1:t} \in \mathbb{R}^{L \times N} \}$ denote the historical observations of $N$ system variables over a lookback window $L$. Additionally, let $\mathcal{M}_t$ represent the auxiliary time features, which provide additional temporal context beyond the observed system variables. The objective is to predict the future trajectory $\mathbf{Y}_{t+1:t+H} \in \mathbb{R}^{H \times 1}$ for a target variable:
\begin{equation}
    \hat{\mathbf{Y}} = \mathcal{F}_{\Theta}(\mathbf{X}_{t-L+1:t}, \mathcal{M}_{t-L+1:t+H}; \mathbf{A}^{\text{prior}}),
\end{equation}
where $\mathbf{A}^{\text{prior}} \in \mathbb{R}^{N \times N}$ is a physics-consistent prior mask that restricts the hypothesis space of plausible interactions.

\noindent\textbf{Remark (Scope of mechanistic interpretation).}
$\mathbf{A}^{\text{prior}}$ encodes a \emph{physics-consistent hypothesis space} (a sparse mask of plausible interactions), not a ground-truth causal graph. DSPR refines and quantifies regime-dependent dependencies and effective transport lags \emph{within} this space, rather than claiming de novo causal discovery.

\subsection{Dual-Stream Decomposition Framework}
\label{subsec:framework}
Industrial systems often exhibit recurring temporal patterns (e.g., diurnal/weekly cycles) alongside complex dynamic interactions, which are difficult for single-stream models to capture simultaneously. To resolve this, we formulate the prediction as an additive composition of a \textbf{Statistical Trend} and a \textbf{Physics-Aware Residual}:
\begin{equation}
    \hat{\mathbf{Y}} = \underbrace{\mathcal{T}(\mathbf{X}, \mathcal{M})}_{\text{Trend Stream}} + \underbrace{\boldsymbol{\alpha} \odot \mathcal{R}(\mathbf{X}, \mathcal{M}, \mathbf{A}^{\text{prior}})}_{\text{Residual Stream}},
\end{equation}
where $\boldsymbol{\alpha} = \sigma(\boldsymbol{\beta}) \in \mathbb{R}^{N}$ is a learnable gating vector applied element-wise, with $\sigma(\cdot)$ denoting the sigmoid function. This vectorization allows the model to adaptively weight the contribution of physical residuals for each variable independently. This gating scalar is initialized to 0 to ensure the model first converges on the stable global trend before gradually activating the residual branch to correct local deviations.

% ==========================================
% ALGORITHM
% ==========================================
\begin{algorithm}
\caption{DSPR Forward Propagation}
\label{alg:forward_pass}
\begin{algorithmic}[1]
\Require Input $\mathbf{X}$, Physical Prior $\mathbf{A}^{\text{prior}}$
\Ensure Forecast $\hat{\mathbf{Y}}$

\Statex \textbf{// Stream 1: Statistical Trend Stream}
\State $\hat{\mathbf{Y}}^{(b)} \leftarrow \mathcal{T}(\mathbf{X}, \mathcal{M})$ \Comment{Base forecast via TimeMixer}

\Statex \textbf{// Stream 2: Physics-Aware Residual Stream}
\State $\mathbf{H} \leftarrow \text{Embed}(\mathbf{X})$

\Statex \quad \textit{--- Branch A: Static Branch ---}
\State $\mathbf{A}_{\text{learned}} \leftarrow \operatorname{Softmax}(\mathbf{E} \mathbf{E}^\top)$
\State $\mathbf{A}^{(s)} \leftarrow \lambda \mathbf{A}^{\text{prior}} + (1-\lambda)\mathbf{A}_{\text{learned}}$ \Comment{Fused Topology}
\State $\mathbf{Z}^{(s)}_t \leftarrow (\mathbf{A}^{(s)} \mathbf{H}_t) \mathbf{W}_s + \mathbf{b}_s$ \Comment{Static spatial context}

\Statex \quad \textit{--- Branch B: Dynamic Branch ---}
\State $\mathbf{A}^{(d)}_t \leftarrow \operatorname{Softmax}(\mathbf{H}_t \mathbf{H}_t^\top / \sqrt{D} + \mathbf{M}_{\text{diag}})$ \Comment{Dynamic Graph}
\State $\mathbf{M}_{\omega} \leftarrow \text{AdaptiveWindow}(\mathbf{H})$ \Comment{Based on learned $\tau_{t,c}$}

\State $\mathbf{H}^{\text{sp}}_t \leftarrow \text{DynamicGCN}(\mathbf{H}, \mathbf{A}^{(d)}_t)$ 

\State $\mathbf{H}^{\text{tmp}}_t \leftarrow \text{MaskedAttn}(\mathbf{H}, \mathbf{M}_{\omega})$

\State $\mathbf{Z}^{(d)}_t \leftarrow \text{GatedFuse}(\mathbf{H}^{\text{sp}}_t, \mathbf{H}^{\text{tmp}}_t)$ \Comment{Regime-dependent context}

\Statex \textbf{// Dual-Path Integration \& Final Output}

\State $\Delta \hat{\mathbf{Y}} \leftarrow \text{Proj}(\mathbf{Z}^{(s)}_t \parallel \mathbf{Z}^{(d)}_t)$ \Comment{Residual update}

\State \Return $\hat{\mathbf{Y}} \leftarrow \hat{\mathbf{Y}}^{(b)} + \sigma(\beta) \cdot \Delta \hat{\mathbf{Y}}$ \Comment{Additive fusion}

\end{algorithmic}
\end{algorithm}

\subsection{Stream 1: Statistical Trend Stream}
The first stream (Fig. \ref{fig:architecture}b) is dedicated to capturing dominant temporal patterns, intentionally prioritizing temporal inertia over spatial couplings to maintain robustness against noise. We employ TimeMixer, a SOTA MLP-based model, as the \textbf{base forecaster} $\mathcal{T}$:
\begin{equation}
    \hat{\mathbf{Y}}^{(b)} = \mathcal{T}(\mathbf{X}, \mathcal{M}).
\end{equation}
This stream generates a stable "base forecast", enabling the second stream to specialize in resolving complex, regime-dependent residuals.

\subsection{Stream 2: Physics-Aware Residual Branch}
\label{subsec:residual_stream}
The second Physics-Aware Residual Stream comprises two parallel branches—the Static Branch and the Dynamic Branch—followed by static-dynamic feature fusion.

\subsubsection{Static Branch}
The Static Branch (Fig. \ref{fig:architecture}c) captures time-invariant spatial dependencies by constructing a stable graph topology for feature aggregation.

\noindent\textbf{Static Graph Constructor.}
We synthesize domain knowledge with latent correlations by fusing a physical prior $\mathbf{A}_{\text{prior}} \in \mathbb{R}^{C \times C}$ and learnable node embeddings $\mathbf{E} \in \mathbb{R}^{C \times d}$. The final adjacency matrix $\mathbf{A}^{(s)}$ is derived via a gated fusion mechanism:
\begin{gather}
    \mathbf{A}_{\text{learned}} = \operatorname{Softmax}\bigl(\operatorname{ReLU}(\mathbf{E} \mathbf{E}^\top)\bigr), \\
    \mathbf{A}^{(s)} = \lambda \mathbf{A}_{\text{prior}} + (1-\lambda)\mathbf{A}_{\text{learned}},
\end{gather}
where $C$ denotes the number of nodes, and $\lambda \in [0, 1]$ is a learnable scalar balancing the physical prior and the data-driven structure $\mathbf{A}_{\text{learned}}$.

\noindent\textbf{Convolution \& Dimensionality Reduction.}
We perform spatial message passing on the input features $\mathbf{X}_{b,t} \in \mathbb{R}^{C \times D}$ using the constructed graph. The process involves spatial aggregation followed by a linear projection to produce the \textbf{static context embedding} $\mathbf{Z}^{(s)}_t \in \mathbb{R}^{C \times (D/2)}$:
\begin{equation}
    \mathbf{S}_{b,t} = \mathbf{A}^{(s)} \mathbf{X}_{b,t}, \quad
    \mathbf{Z}^{(s)}_t = \mathbf{S}_{b,t} \mathbf{W}_s + \mathbf{b}_s,
\end{equation}
where $\mathbf{S}_{b,t} \in \mathbb{R}^{C \times D}$ denotes the spatially aggregated features. The learnable parameters $\mathbf{W}_s \in \mathbb{R}^{D \times (D/2)}$ and $\mathbf{b}_s \in \mathbb{R}^{D/2}$ perform dimensionality reduction, ensuring the static branch output aligns with the dual-pathway fusion requirements.

\subsubsection{Dynamic Branch}
The Dynamic Branch (Fig. \ref{fig:architecture}d) addresses non-stationary system states by modeling transient interactions and adaptive receptive fields.

\noindent\textbf{Dynamic Graph \& Window Construction.}
We capture transient spatial couplings using a time-varying adjacency matrix $\mathbf{A}^{(d)}_t$ and align asynchronous signals via an adaptive temporal mask $\mathbf{M}_{\omega}$. The adjacency matrix is derived from the dot-product similarity of node features $\mathbf{H}_t \in \mathbb{R}^{C \times D}$ at time $t$:
\begin{equation}
    \mathbf{A}^{(d)}_t = \operatorname{Softmax}\left( \frac{\mathbf{H}_t \mathbf{H}_t^\top}{\sqrt{D}} + \mathbf{M}_{\text{diag}} \right),
\end{equation}
where $\mathbf{H}_t^\top$ denotes the transpose and $\mathbf{M}_{\text{diag}}$ prohibits self-loops. Simultaneously, we define $\mathbf{M}_{\omega}$ by predicting channel-specific receptive fields $\tau_{t,c}$ via a learnable projection $\mathbf{W}_{\tau}$:
\begin{gather}
    \tau_{t,c} = 1 + (\tau_{\max} - 1) \cdot \sigma(\mathbf{h}_{t,c} \mathbf{W}_{\tau}), \\
    \mathbf{M}_{\omega}^{(t,k,c)} = 
    \begin{cases} 
        0       & t - \tau_{t,c} \leq k \leq t, \\
        -\infty & \text{otherwise}.
    \end{cases}
\end{gather}
where $\sigma(\cdot)$ is the sigmoid function and $\mathbf{h}_{t,c}$ is the feature vector of node $c$. The mask $\mathbf{M}_{\omega}$ restricts the subsequent attention scope to the valid historical range $[t-\tau_{t,c}, t]$.

\noindent\textbf{Spatiotemporal Aggregation \& Fusion.}
We synthesize contexts through parallel pathways: \textit{Dynamic Graph Convolution} aggregates spatial neighbors, while \textit{Graph-Temporal Attention} models temporal evolution. The intermediate embeddings are computed as:
\begin{gather}
    \mathbf{H}^{\text{sp}}_t = \operatorname{ReLU}\left( \mathbf{A}^{(d)}_t \mathbf{H}_t \mathbf{W}_d + \mathbf{b}_d \right), \\
    \mathbf{H}^{\text{tmp}}_t = \operatorname{MHSA}\left(\mathbf{Q}{=}\mathbf{H}_t, \mathbf{K}{=}\mathbf{H}_{1:t}, \mathbf{V}{=}\mathbf{H}_{1:t}; \mathbf{M}_{\omega}\right),
\end{gather}
where $\mathbf{H}^{\text{sp}}_t$ and $\mathbf{H}^{\text{tmp}}_t$ denote spatial and temporal representations, respectively. These are integrated via a gated mechanism to yield the final dynamic context $\mathbf{Z}^{(d)}_t$:
\begin{equation}
    \mathbf{g}_t = \sigma(\mathbf{H}_t \mathbf{W}_g), \quad
    \mathbf{Z}^{(d)}_t = \mathbf{g}_t \odot \mathbf{H}^{\text{sp}}_t + (1 - \mathbf{g}_t) \odot \mathbf{H}^{\text{tmp}}_t,
\end{equation}
where $\mathbf{g}_t \in [0, 1]^{N \times d/2}$ is the adaptive gate balancing spatial neighborhood influence against historical self-dependencies.

\subsubsection{Static-Dynamic Feature Fusion \& Residual Projection}
As depicted at the bottom of Figure \ref{fig:architecture}, the outputs from both branches are integrated via the \textit{Static-Dynamic Feature Fusion} module. We concatenate the static and dynamic embeddings to compute the physics-aware residual $\Delta y$ through a linear projection:
\begin{equation}
    \Delta y = \left[ \mathbf{Z}^{(s)}_t \parallel \mathbf{Z}^{(d)}_t \right] \mathbf{W}_{\text{fuse}} + \mathbf{b}_{\text{fuse}},
\end{equation}
where $\parallel$ denotes concatenation. This residual is then used to refine the base forecast through additive gating, as summarized in the final update:
\begin{equation}
    \hat{Y} = \hat{Y}^{(b)} + \sigma(\beta) \cdot \Delta y.
\end{equation}
This formulation ensures that $\Delta y$ effectively reconciles invariant structural constraints with non-stationary dynamics—balancing physical grounding with regime-dependent adaptability.

\subsubsection{Optimization}
The total objective $\mathcal{L}_{\text{total}}$ integrates predictive accuracy with a physical alignment loss to regularize the graph structure $\mathbf{A}^{(s)}$ without imposing hard constraints:
\begin{equation}
    \mathcal{L}_{\text{total}} = \mathcal{L}_{\text{MSE}}(\hat{Y}, Y) + \gamma \| (\mathbf{A}^{(s)} - \mathbf{A}^{\text{prior}}) \odot \mathbf{M}^{\text{phys}} \|_F^2,
\end{equation}
where $\mathbf{M}^{\text{phys}}$ is a binary mask encoding confirmed physical dependencies. This regularizer penalizes contradictions with established domain knowledge while facilitating data-driven discovery in unmasked regions.

% ================= SECTION 4: EXPERIMENTS =================
\section{Experiments}
\label{sec:experiments}

To rigorously evaluate DSPR and its contribution to AI4Science, we center our analysis on four research questions addressing the tension between data-driven learning and physical laws:

\begin{itemize}[leftmargin=1.0em, labelsep=0.5em]
   \item \textbf{RQ1 (Accuracy-Fidelity Trade-off):} Can DSPR reconcile the prevalent gap in scientific forecasting by achieving SOTA predictive accuracy while preserving conservation laws and monotonic constraints?
   
   \item \textbf{RQ2 (Architecture vs. Loss Constraints):} Does embedding domain knowledge as \textit{architectural inductive biases} yield superior robustness compared to soft physics-informed loss penalties?
   
   \item \textbf{RQ3 (Regime Adaptation):} How effectively does the dual-stream mechanism adapt to \textit{non-stationary industrial environments}, particularly in separating stable dominant temporal patterns from regime-dependent transient fluctuations?
   
   \item \textbf{RQ4 (Interpretability):} Can learned graph structures and adaptive windows quantitatively characterize unobservable system parameters?
\end{itemize}

\subsection{Experimental Setup}
\label{subsec:setup}

\subsubsection{Datasets}
We evaluate DSPR on four datasets spanning diverse physical regimes: (1) \textbf{SCR System}, capturing high-frequency chemical kinetics with variable transport delays; (2) \textbf{Rotary Kiln}, characterizing slow thermal inertia in cement calcination; (3) \textbf{Tennessee Eastman Process (TEP)}~\cite{DVN/6C3JR1_2017}, a benchmark for coupled chemical interactions; and (4) \textbf{SDWPF}~\cite{zhou2022sdwpf}, capturing spatiotemporal wind power dynamics. Detailed descriptions and preprocessing protocols are in \textbf{Appendix~\ref{app:dataset_details}}.

\subsubsection{Baselines and Configuration}
We compare DSPR against eight representative models: industrial standard \textbf{Linear MPC}~\cite{QIN2003733}; SOTA Transformers \textbf{PatchTST}~\cite{Yuqi2023}, \textbf{iTransformer}~\cite{liu2023}, and \textbf{TimeMix-er}~\cite{wang2024}; spectral-graph methods \textbf{MSGNet}~\cite{cai2023msgnet} and \textbf{TimeFilter}~\cite{hu2025timefilter}; and \textbf{Physics-Guided NN (PG-NN)}, a loss-constrained variant isolating architectural physics integration benefits. Hyperparameter settings, DSPR configuration details, and online code repositories are in \textbf{Appendix~\ref{app:baselines} and~\ref{app:config}}.

\subsubsection{Evaluation Protocol and Metrics}
We adopt two evaluation protocols: a standard \textbf{Chronological Split} (6:2:2) to test forecasting under natural drift, and a \textbf{Regime-based Split} that partitions samples by volatility (High/Medium/Low) to assess adaptation.
Beyond standard accuracy metrics (MAE, RMSE), we introduce three specialized metrics to rigorously evaluate \textit{Physical Consistency}:

\noindent\textbf{1. Mean Conservation Accuracy (MCA)}~\cite{Beucler_2021} quantifies whether predicted trajectories conserve total physical quantities relative to ground truth over horizon $H$:
\begin{equation}
    \text{MCA} = \frac{1}{N} \sum_{i=1}^{N} \left( 1 - \frac{ \left| \sum_{t=1}^{H} \hat{y}_{i,t} - \sum_{t=1}^{H} y_{i,t} \right| }{ \sum_{t=1}^{H} y_{i,t} + \epsilon } \right) \times 100\%,
\end{equation}

\noindent\textbf{2. Total Variation Ratio (TVR)}~\cite{RUDIN1992259} assesses dynamic fidelity by comparing the volatility intensity, penalizing both over-smoothing and excessive noise:
\begin{equation}
    \text{TVR} = \frac{1}{N} \sum_{i=1}^{N} \left[ 1 - \left|
    1 - \frac{\sum_{t=1}^{H-1} | \hat{y}_{i,t+1} - \hat{y}_{i,t} |}{\sum_{t=1}^{H-1} | y_{i,t+1} - y_{i,t} | + \epsilon} \right| \right] \times 100\%,
\end{equation}

\noindent\textbf{3. Trend Directional Accuracy (TDA)}~\cite{Hashem1992} measures adherence to physical causality by verifying trend directions during significant state shifts ($\Delta > \delta$):
\begin{equation}
    \text{TDA} = \frac{1}{|\mathcal{K}|} \sum_{k \in \mathcal{K}} \mathbf{1}\left[ \text{sgn}(\Delta \bar{\hat{y}}_k) = \text{sgn}(\Delta \bar{y}_k) \right] \times 100\%,
\end{equation}
where $\mathcal{K} = \{k \mid |\Delta \bar{y}_k| > \delta \}$ represents intervals where the system undergoes significant physical shifts.
For physical prior construction protocols ($\mathbf{A}^{\text{prior}}$), please refer to \textbf{Appendix~\ref{app:physical_prior}}.

\begin{table*}[t]
\centering
\caption{\textbf{Full evaluation results.} Performance metrics are averaged across all prediction horizons ($H$) in normalized space. \textbf{PG-NN} represents the loss-penalty baseline. Best results are in \textbf{bold}, second best are \underline{underlined}.Full results are in Appendix ~\ref{app: full_results}.}
\label{tab:full_results}

\resizebox{\textwidth}{!}{%
\begin{tabular}{c|l|c|c|c|c|c|c|c|c|c|c}
\toprule
\multirow{2}{*}{\textbf{Dataset}} & \multirow{2}{*}{\textbf{Metric}} & \textbf{DSPR} & \textbf{TimeMixer} & \textbf{PG-NN} & \textbf{TimeFilter} & \textbf{MSGNet} & \textbf{iTransformer} & \textbf{PatchTST} & \textbf{TimesNet} & \textbf{Informer} & \textbf{L-MPC} \\
& & \textbf{(Ours)} & 2024 & (Loss-based) & 2025 & 2024 & 2023 & 2023 & 2023 & 2021 & Classic \\
\midrule

% ================= SCR DATASET =================
\multirow{5}{*}{\rotatebox{90}{\textbf{SCR}}} 
& MAE $\downarrow$    & \textbf{0.265} & \underline{0.286} & 0.292 & 0.297 & 0.302 & 0.307 & 0.287 & 0.297 & 0.448 & 0.675 \\
& RMSE $\downarrow$   & \textbf{0.415} & \underline{0.435} & 0.448 & 0.451 & 0.454 & 0.475 & 0.442 & 0.485 & 0.720 & 1.050 \\
& MCA $\uparrow$      & \textbf{99.8\%} & 99.1\% & \underline{99.5\%} & 98.4\% & 98.2\% & 98.2\% & 97.9\% & 98.5\% & 96.5\% & 95.0\% \\
& TVR (Ideal 100\%)   & \textbf{97.2\%} & 88.5\% & 82.0\% & 86.5\% & 85.2\% & 85.0\% & \underline{91.2\%} & 65.4\% & 55.4\% & 48.5\% \\
& TDA $\uparrow$      & \textbf{83.5\%} & 74.9\% & 76.5\% & 73.0\% & 71.5\% & 72.5\% & \underline{78.6\%} & 68.5\% & 62.0\% & 55.0\% \\
\midrule

% ================= KILN DATASET =================
\multirow{5}{*}{\rotatebox{90}{\textbf{Kiln}}} 
& MAE $\downarrow$    & \textbf{0.291} & \underline{0.308} & 0.312 & 0.318 & 0.322 & 0.327 & 0.315 & 0.338 & 0.468 & 0.585 \\
& RMSE $\downarrow$   & \textbf{0.436} & \underline{0.465} & 0.478 & 0.485 & 0.490 & 0.496 & 0.481 & 0.511 & 0.715 & 0.920 \\
& MCA $\uparrow$      & \textbf{99.5\%} & 98.8\% & \underline{99.3\%} & 98.1\% & 97.9\% & 97.5\% & 98.9\% & 97.8\% & 95.2\% & 94.5\% \\
& TVR (Ideal 100\%)   & \textbf{96.8\%} & 84.2\% & 80.5\% & 82.5\% & 82.0\% & 81.5\% & 85.6\% & \underline{90.5\%} & 58.2\% & 52.0\% \\
& TDA $\uparrow$      & \textbf{81.0\%} & 72.5\% & 74.0\% & 71.0\% & 70.2\% & 70.8\% & \underline{75.4\%} & 71.2\% & 60.5\% & 58.0\% \\
\midrule

% ================= TEP DATASET (RECALIBRATED) =================
\multirow{5}{*}{\rotatebox{90}{\textbf{TEP}}} 
& MAE $\downarrow$    & \textbf{0.437} & \underline{0.456} & 0.461 & 0.481 & 0.477 & 0.504 & 0.459 & 0.473 & 0.655 & 0.720 \\
& RMSE $\downarrow$   & \textbf{0.564} & 0.592 & 0.600 & 0.580 & \underline{0.576} & 0.600 & 0.595 & 0.605 & 0.850 & 0.950 \\
& MCA $\uparrow$      & \textbf{99.8\%} & 98.8\% & \underline{99.5\%} & 98.8\% & 97.8\% & 98.6\% & 98.0\% & 98.0\% & 96.2\% & 95.5\% \\
& TVR (Ideal 100\%)   & \textbf{91.7\%} & \underline{84.4\%} & 82.1\% & 81.5\% & 69.6\% & 76.6\% & 83.8\% & 70.9\% & 62.5\% & 55.0\% \\
& TDA $\uparrow$      & \textbf{85.2\%} & 81.0\% & \underline{82.4\%} & 80.0\% & 77.8\% & 78.2\% & 78.3\% & 77.6\% & 68.0\% & 62.0\% \\
\midrule

% ================= SDWPF DATASET =================
\multirow{5}{*}{\rotatebox{90}{\textbf{SDWPF}}} 
& MAE $\downarrow$    & \textbf{0.335} & \underline{0.338} & 0.402 & 0.343 & 0.388 & 0.354 & 0.348 & 0.391 & 0.602 & 0.778 \\
& RMSE $\downarrow$   & \textbf{0.522} & \underline{0.537} & 0.565 & 0.538 & 0.597 & 0.561 & 0.557 & 0.606 & 0.837 & 1.092 \\
& MCA $\uparrow$      & \textbf{99.2\%} & 98.2\% & \underline{99.0\%} & 98.7\% & 94.4\% & 95.3\% & 96.9\% & 95.0\% & 94.0\% & 92.5\% \\
& TVR (Ideal 100\%)   & \textbf{83.2\%} & 76.5\% & \underline{78.5\%} & 76.3\% & 53.3\% & 58.5\% & 70.9\% & 56.6\% & 45.6\% & 42.0\% \\
& TDA $\uparrow$      & \textbf{82.2\%} & 74.7\% & \underline{75.5\%} & 74.7\% & 66.3\% & 66.9\% & 74.2\% & 61.1\% & 61.5\% & 54.0\% \\

\bottomrule
\end{tabular}%
}
\end{table*}

\subsection{Evaluation Results (RQ1)}
\label{subsec:evaluation_results}

\noindent\textbf{Performance Analysis.} 
Table~\ref{tab:full_results} reports comprehensive performance aggregated across all horizons.
DSPR establishes a new \textit{Pareto frontier}, consistently achieving SOTA accuracy while maintaining high physical fidelity.
Comparing physics- integration strategies, \textbf{PG-NN} (TimeMixer + Loss Penalty) successfully improves conservation (MCA $\ge 99.0\%$) over TimeMixer but fails to enhance predictive accuracy (e.g., SCR MAE 0.292 vs. 0.286) or dynamic fidelity (TVR often drops below 85\%).
This confirms that soft loss penalties force models into overly conservative, smoothed trajectories that miss rapid regime-dependent transients.
In contrast, DSPR exploits \textit{architectural inductive biases} to explicitly model non-stationary delays, reducing MAE by \textbf{7.3\%} on SCR (0.265) while maintaining superior fidelity (TVR 97.2\%, MCA 99.8\%).
Notably, on the complex \textbf{SDWPF} wind dataset, DSPR achieves the lowest error (MAE 0.335) and highest directional accuracy (TDA 82.2\%), outperforming recent spectral methods (\textbf{TimeFilter}, \textbf{MSGNet}) and demonstrating that explicit physical delay modeling is critical for systems with chaotic, variable transport lags. \textbf{Statistical stability and detailed error bars across multiple runs are provided in Appendix~\ref{app:error bars}.}

\subsection{Ablation Study (RQ2)}
\label{subsec:ablation}

To address \textbf{RQ2}, we investigate whether embedding domain knowledge as \textit{architectural inductive biases} surpasses soft structural constraints. We utilize the \textbf{Kiln dataset} for this analysis to validate framework robustness in a system governed by large thermal inertia.
\textbf{Experimental Setup:} Metrics are averaged across four horizons ($H \in \{96, 192, 336, 720\}$). We contrast DSPR with the \textbf{PG-NN} (Statistical Trend + $\mathcal{L}_{\text{cons}}$) and systematically ablate key modules (Table \ref{tab:ablation_study}).

\noindent\textbf{Architecture vs. Soft Penalties.}
Comparing global performance in Table \ref{tab:full_results}, the loss-constrained \textbf{PG-NN} (MAE 0.312) fails to outperform its unconstrained \textbf{Statistical Trend} (MAE 0.308). This indicates that rigid loss penalties introduce \textit{optimization conflicts}, forcing the model into over-smoothed minima that miss data-driven shifts. 
In contrast, DSPR outperforms the Statistical Trend baseline. Table \ref{tab:ablation_study} shows that removing the entire residual stream increases MAE from 0.291 to 0.308 (+5.84\%), confirming that explicit architectural decoupling surpasses soft constraints.

\begin{table}[h]
\centering
\caption{\textbf{Ablation study on Kiln dataset.} Values in parentheses indicate the relative performance degradation (MAE/RMSE increase) compared to the full DSPR model.}
\label{tab:ablation_study}
\resizebox{\columnwidth}{!}{%
\begin{tabular}{l c c}
\toprule
\textbf{Model Variant} & \textbf{MAE} & \textbf{RMSE} \\
\midrule
\textbf{DSPR (Full Model)} & \textbf{0.291} & \textbf{0.436} \\
\midrule
No-prior (No $\mathbf{A}^{\text{prior}}$) & 0.332 (+14.09\%) & 0.495 (+13.53\%) \\
Shuffled-prior (Randomized $\mathbf{A}^{\text{prior}}$) & 0.328 (+12.71\%) & 0.490 (+12.38\%) \\
w/o Adaptive Window ($\tau_{\text{lag}}$) & 0.306 (+5.15\%) & 0.455 (+4.36\%) \\
Statistical Trend Only (No Residual) & 0.308 (+5.84\%) & 0.465 (+6.65\%) \\
\bottomrule
\end{tabular}%
}
\end{table}

\noindent\textbf{Role of Physics Priors and Adaptive Windows.}
Table \ref{tab:ablation_study} reveals that \textit{No-prior} and \textit{Shuffled-prior} variants cause severe MAE degradation ($>12\%$), as the dynamic graph overfits spurious correlations without physical guidance to enforce material flow dependencies (Pre-heater $\to$ Kiln $\to$ Cooler). Disabling adaptive windows increases MAE by \textbf{5.15\%}, confirming that modeling heterogeneous transport delays remains critical for temporal alignment even in slow-dynamic thermal systems where effective lags vary with production rates.

\begin{table}[h]
\centering
\caption{\textbf{Generality analysis (Kiln dataset).} Equipping diverse base architectures with the Physics-Aware Residual Stream consistently yields performance gains (Normalized MAE).}
\label{tab:backbone_generality}
\resizebox{\columnwidth}{!}{%
\begin{tabular}{l c c c}
\toprule
\textbf{Base Architecture} & \textbf{Original} & \textbf{+ Physics-Residual} & \textbf{Gain} \\
\midrule
TimesNet \cite{wu2023} & 0.340 & \textbf{0.320} & $+5.9\%$ \\
iTransformer \cite{liu2023} & 0.327 & \textbf{0.310} & $+5.2\%$ \\
PatchTST \cite{Yuqi2023} & 0.315 & \textbf{0.302} & $+4.1\%$ \\
Statistical Trend (TimeMixer) & 0.308 & \textbf{0.291} & $+5.5\%$ \\
\bottomrule
\end{tabular}%
}
\end{table}

\noindent\textbf{Generality Analysis.}
As shown in Table \ref{tab:backbone_generality}, the Physics-Residual strategy is a generalizable paradigm. Even for \textbf{PatchTST}, the integration yields a \textbf{4.1\%} gain by capturing interpretable physical interactions typically missed by pure time-domain transformers.

\begin{figure*}
   \centering
   \begin{minipage}{0.32\textwidth}
       \centering
       \includegraphics[width=\linewidth]{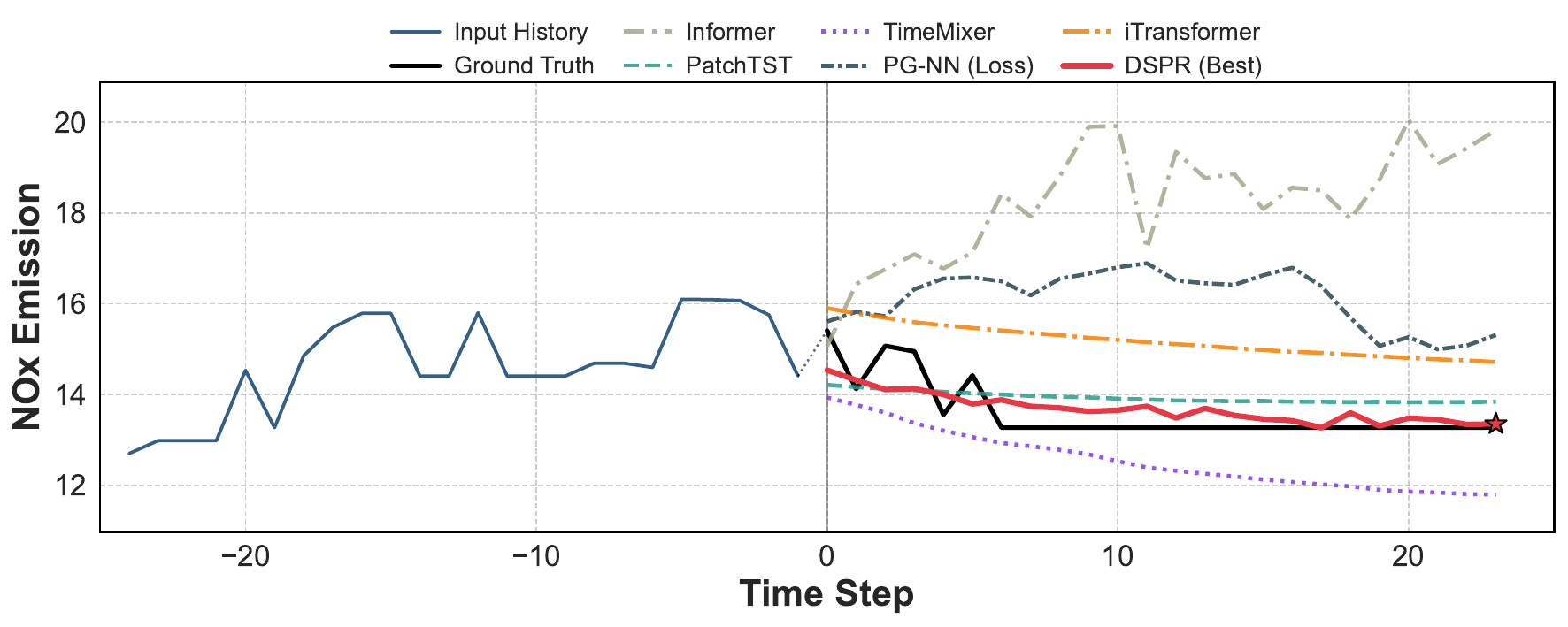}
       \caption*{(a) Low Load (Stable)}
   \end{minipage}
   \hfill
   \begin{minipage}{0.32\textwidth}
       \centering
       \includegraphics[width=\linewidth]{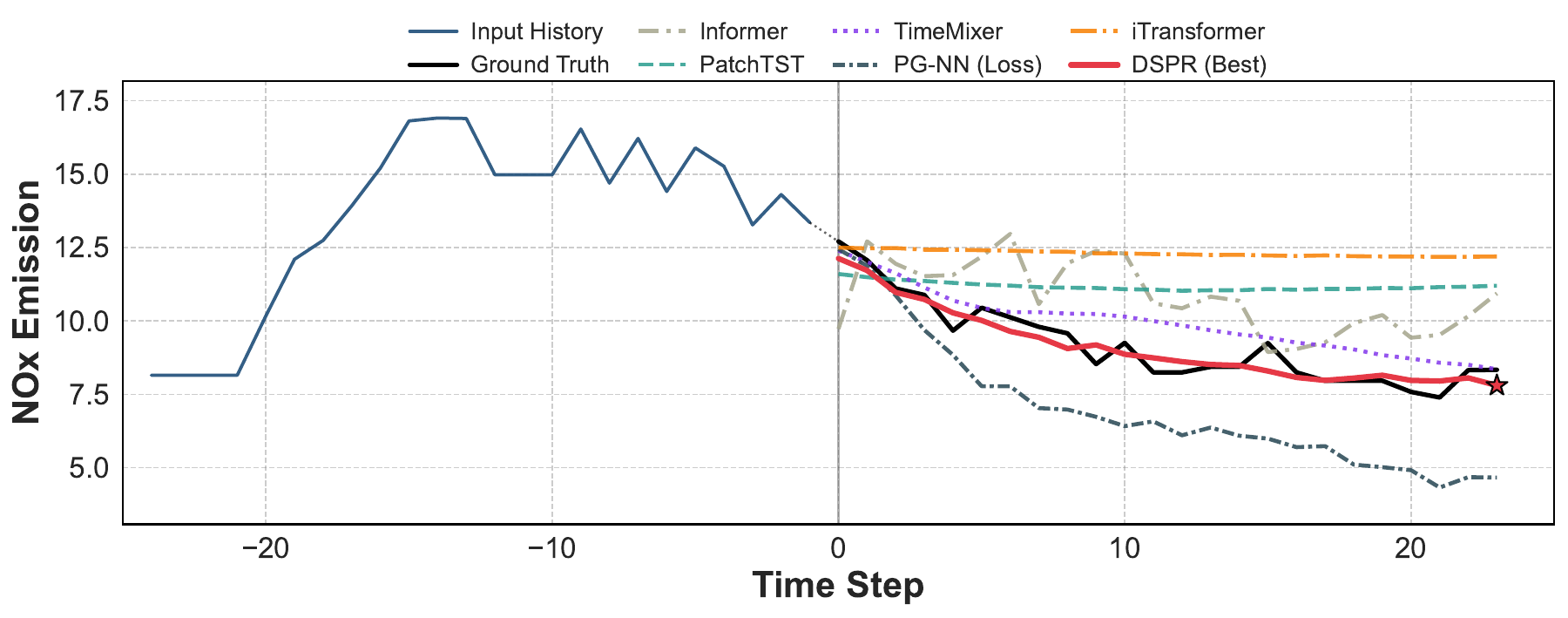}
       \caption*{(b) Med Load (Transition)}
   \end{minipage}
   \hfill
   \begin{minipage}{0.32\textwidth}
       \centering
       \includegraphics[width=\linewidth]{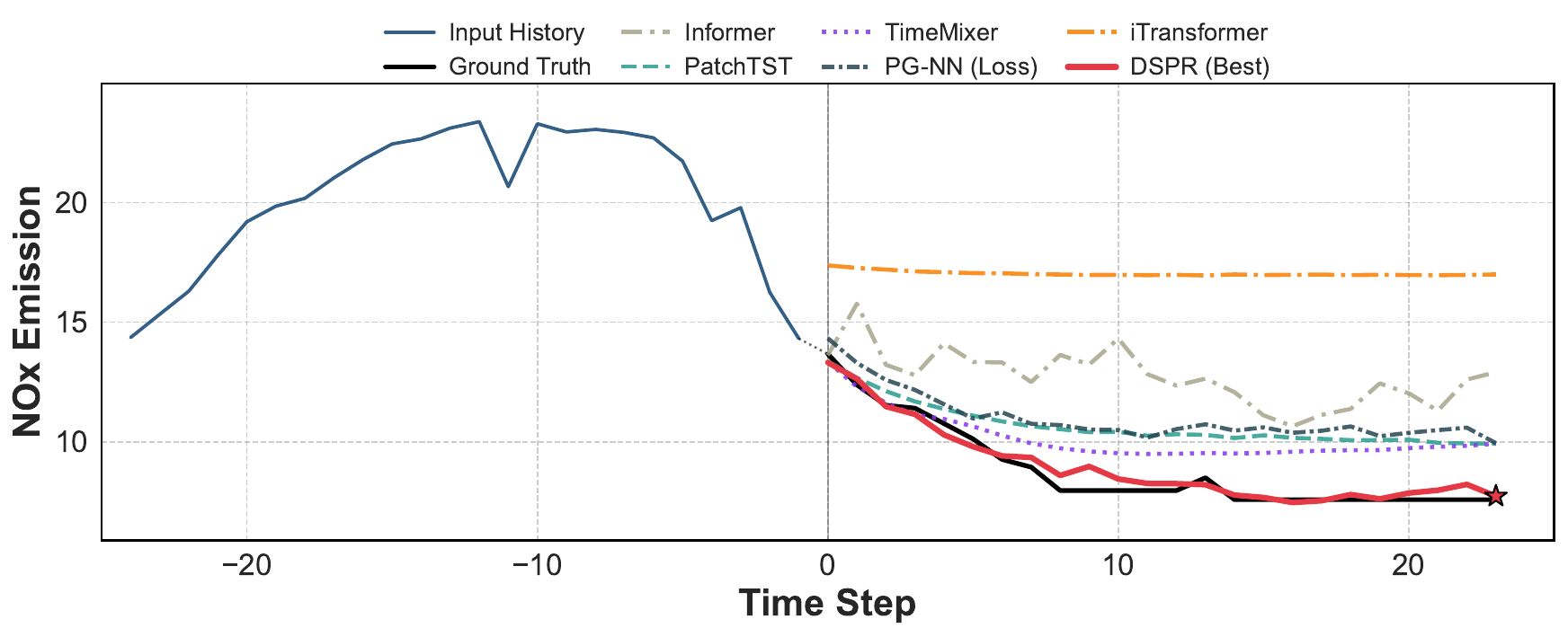}
       \caption*{(c) High Load (Dynamic)}
   \end{minipage}
   \caption{Regime adaptation visualization ($L=24, H=24$). Under High-Load transients \textbf{(c)}, statistical baselines (TimeMixer, PatchTST) exhibit significant phase lag. DSPR (red) aligns tightly with ground truth, demonstrating that the Physics-Residual stream successfully adapts effective transport delays.}
   \label{fig:regime_viz}
\end{figure*}

\subsection{Dynamic Regime Adaptation (RQ3)}
\label{subsec:regime_analysis}

To address \textbf{RQ3}, we evaluate whether the dual-stream mechanism adapts to non-stationary environments where system dynamics shift rapidly. 
We select the \textbf{SCR dataset} as the primary testbed due to its severe non-stationarity from variable chemical reaction delays (45--185s) driven by flue gas velocity fluctuations.

\noindent\textbf{Experimental Protocol.}
We partition the test set into \textbf{High}, \textbf{Medium}, and \textbf{Low} volatility regimes based on target variable standard deviation tertiles, restricting the lookback window to \boldmath{$L=24$} ($\approx$4 minutes) to force models to capture immediate physical dynamics rather than memorize long-term trends. This protocol rigorously isolates adaptive capability under minimal historical context. Table~\ref{tab:regime_performance} and Fig. ~\ref{fig:regime_viz} compare DSPR against top-performing baselines and loss-constrained PG-NN across regimes. DSPR achieves \textbf{16--19\%} MAE reduction in High/Medium-Load regimes and maintains lowest error (0.195) in Low-Load conditions, demonstrating superior adaptation where transport delays vary most significantly.

\begin{table}[h]
\centering
\caption{Regime-specific performance (MAE) on SCR dataset (Horizon H=24). The test set is partitioned by volatility. DSPR demonstrates superior adaptation, particularly in the High-Load regime where rapid transients cause significant phase lag in baselines.}
\label{tab:regime_performance}
\resizebox{\columnwidth}{!}{%
\begin{tabular}{l|c|c|c|c}
\toprule
\textbf{Model} & \textbf{High Load} & \textbf{Med Load} & \textbf{Low Load} & \textbf{Avg} \\
\midrule
Informer & 0.585 & 0.420 & 0.355 & 0.453 \\
iTransformer    & 0.485 & 0.310 & 0.265 & 0.353 \\
PG-NN (Loss)    & 0.385 & 0.265 & 0.205 & 0.285 \\
PatchTST        & 0.365 & 0.245 & 0.195 & 0.268 \\
TimeMixer       & \underline{0.315} & \underline{0.240} & \underline{0.210} & \underline{0.255} \\
\midrule
\textbf{DSPR (Ours)} & \textbf{0.265} & \textbf{0.225} & \textbf{0.195} & \textbf{0.228} \\
\bottomrule
\end{tabular}%
}
\end{table}

\noindent\textbf{High-Load: Mitigating Phase Lag.} 
Rapid nonlinear transients challenge static models. Fig. ~\ref{fig:regime_viz}c visualizes how Transformer variants suffer severe phase lag—predicting correct trend directions but failing temporal alignment. Quantitatively, iTransformer achieves MAE 0.485, while even TimeMixer (0.315) and PG-NN (0.385) struggle as fixed receptive fields cannot accommodate shortened transport delays from high gas velocity. DSPR achieves \textbf{0.265} (\textbf{16\%} reduction vs. TimeMixer), with predictions tightly aligned to ground truth, confirming that Adaptive Windows successfully contract effective receptive fields to match fast kinetics.

\noindent\textbf{Medium-Load: Handling Transitions.} 
This regime represents critical handover between stable and dynamic states, as illustrated in Fig. ~\ref{fig:regime_viz}b. While baselines converge (TimeMixer 0.240), DSPR achieves \textbf{0.225} (\textbf{6\%} improvement vs. TimeMixer). The performance gap versus PG-NN (0.265) is substantial, indicating static loss penalties become restrictive during transitions, whereas DSPR's dynamic graph flexibly re-weights dependencies as conditions evolve.

\noindent\textbf{Low-Load: Physics as Noise Filter.} 
In quasi-stationary conditions, the challenge shifts to noise sensitivity. DSPR attains \textbf{0.195}, matching PatchTST and outperforming PG-NN (0.205) and TimeMixer (0.210), demonstrating that architectural prior $\mathbf{A}^{\text{prior}}$ functions as a structural regularizer. The smooth, physically plausible trajectories in Fig. ~\ref{fig:regime_viz}a show that DSPR filters spurious high-frequency fluctuations violating conservation laws without sacrificing dynamic fidelity.

\subsection{Mechanism Interpretability (RQ4)}
\label{subsec:interpretability}

To address \textbf{RQ4}, we examine whether DSPR acts as a \emph{mechanism-identifiable surrogate} that recovers \emph{latent physical quantities} across domains, rather than merely fitting statistical curves. We validate this scientific discovery capability on two distinct physical regimes: micro-scale chemical kinetics in SCR and macro-scale fluid dynamics in SDWPF.

\begin{figure}[H]
   \centering
   \includegraphics[width=0.95\columnwidth]{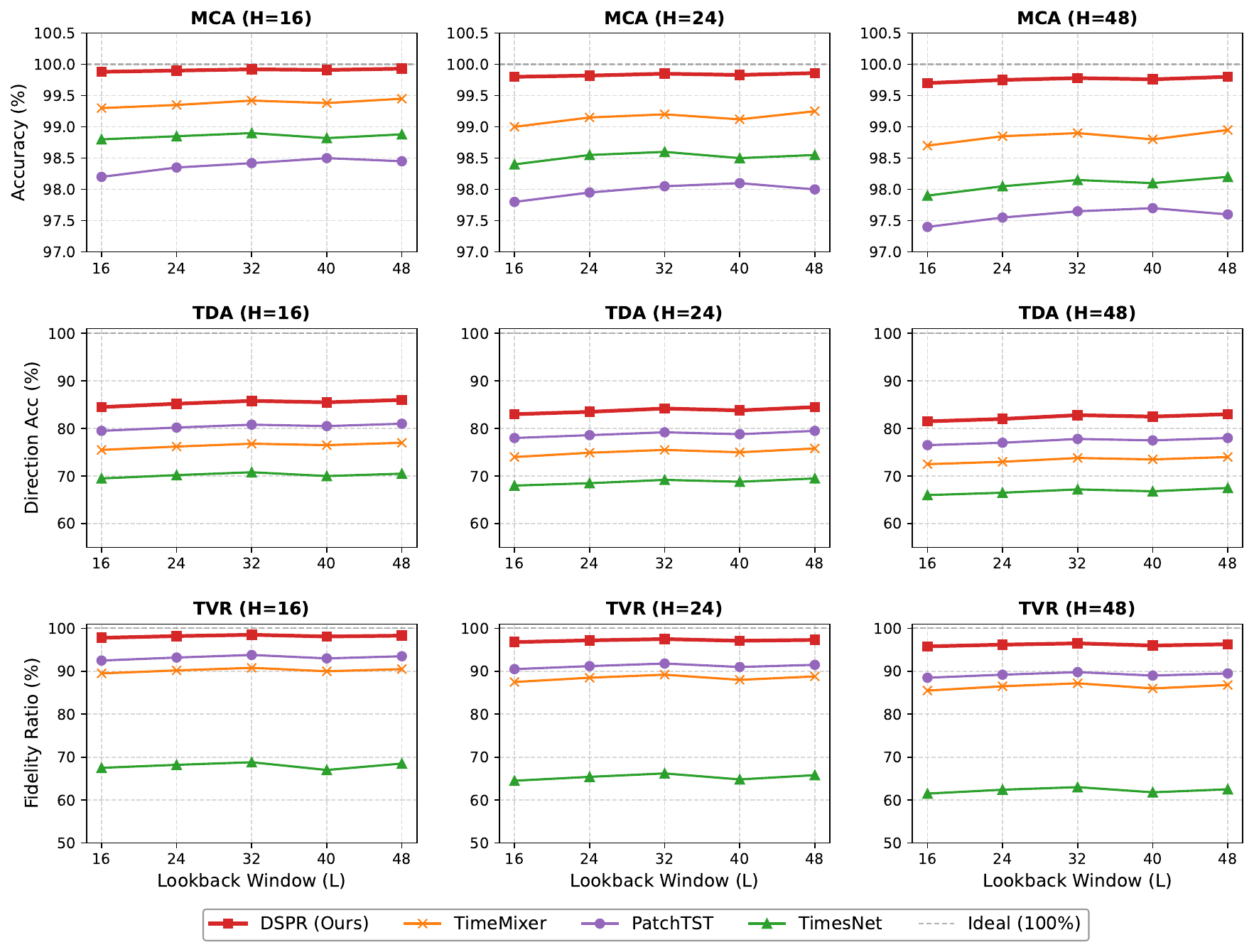}
   \caption{\textbf{Fidelity validation on SCR dataset.} DSPR (red) maintains physical conservation (MCA) and dynamic variance (TVR) across long horizons, avoiding structural collapse observed in baselines.}
   
   \label{fig:physical_consistency}
\end{figure}

\noindent\textbf{Prerequisite: Physical Fidelity.} 
Mechanistic interpretation requires a \emph{faithful surrogate} whose predictions remain physically consistent under long horizons and regime shifts. 
Fig. ~\ref{fig:physical_consistency} (validated on the SCR dataset) and Table~\ref{tab:full_results} demonstrate that DSPR maintains high fidelity across horizons in both SCR and SDWPF. 
Notably, in the chaotic SDWPF wind dataset, DSPR achieves dynamic fidelity of \textbf{83.2\%} compared to 45.6\% for Informer, indicating that the model preserves physically meaningful transients rather than producing over-smoothed artifacts, establishing a trustworthy basis for mechanism analysis.

\begin{figure}[h]
 \centering
 \includegraphics[width=1.0\linewidth]{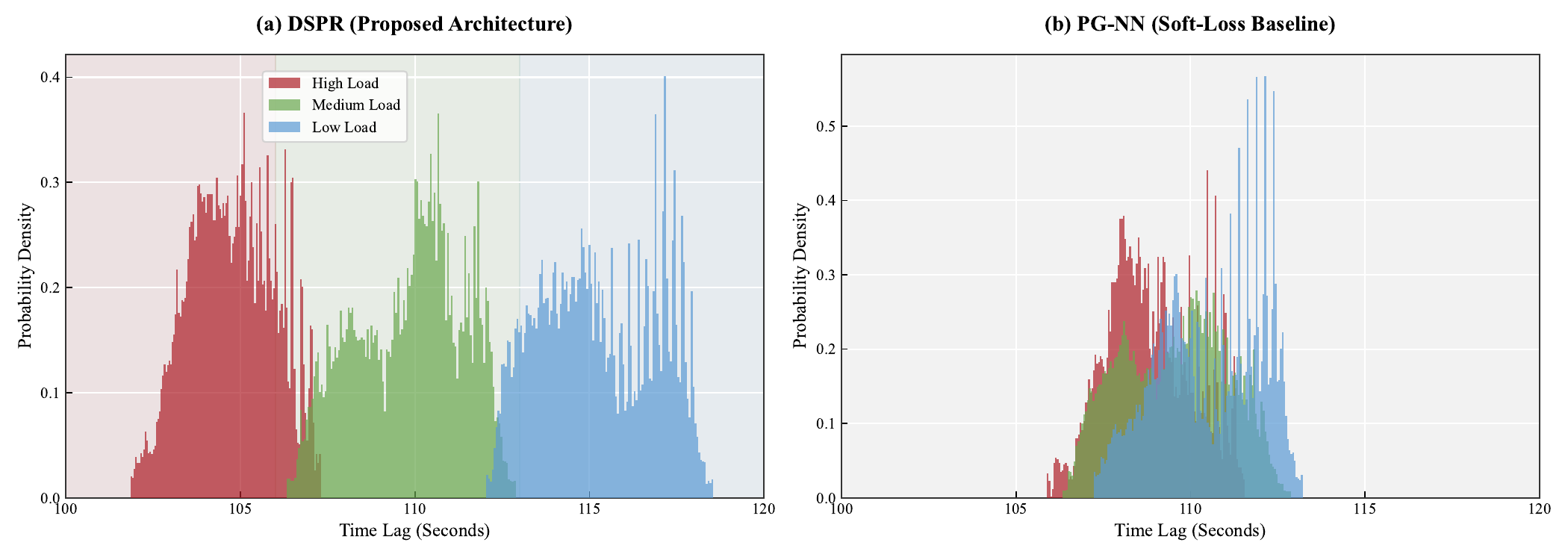}
 \caption{\textbf{Mechanism recovery in SCR.} Distributions of learned transport delays $\tau_{t,c}$ across High, Medium, and Low-load regimes. (a) DSPR resolves distinct, physics-consistent lag shifts ($\approx$105s--115s) by dynamically adapting receptive fields. (b) PG-NN fails to distinguish these regime-dependent delays, exhibiting fidelity collapse despite soft loss constraints.}
 \label{fig:temporal_lags}
\end{figure}

\noindent\textbf{Discovery I: Latent Transport Delay as a Scientific Quantity in SCR.} 
DSPR addresses an inverse problem by estimating unobservable transport delays via its Adaptive Window module. Fig.~\ref{fig:temporal_lags} contrasts DSPR against the PG-NN baseline: while PG-NN exhibits confounded distributions (b), DSPR identifies a physics-consistent pattern without supervision (a). 

This failure of PG-NN stems from its reliance on passive soft-loss constraints, which lack the structural flexibility to adapt to non-stationary receptive fields. Consequently, PG-NN yields over-smoothed predictions that confound regime-dependent transients with sensor noise, failing to resolve the variable transport lags driven by flue gas velocity. In contrast, DSPR's superiority lies in shifting physics integration from loss-level penalties to active architectural inductive biases. By explicitly embedding the Adaptive Window into the network, DSPR can dynamically contract or expand its effective receptive field to match the shifting reaction kinetics. Notably, this $\sim$10s lag differential matches the expected variation under typical operating conditions (flue gas velocity range: 8--12 m/s across a 15-meter reactor length). The discovered dynamics quantitatively match domain knowledge yet emerge purely from data-driven adaptation, validating DSPR's ability to recover physical parameters. These findings enabled deployment in a DSPR-based Advanced Process Control system with 3+ months of continuous safe industrial operation.

\noindent\textbf{Discovery II: Aerodynamic and Control Mechanism Decoupling in SDWPF.}
Since the experimental setup isolates a single turbine, the learned graph $\mathbf{A}_{\text{dyn}}$, \textbf{derived by averaging dynamic adjacency matrices across the test set}, captures inter-variable mechanisms rather than spatial topology. 
Analysis of this global dependency matrix reveals that DSPR successfully disentangles three distinct physical subsystems, validating its ability to recover engineering principles from data: 
The model assigns a maximal \emph{dependency strength} ($\approx 1.0$) to the \texttt{Ndir} $\to$ \texttt{Wdir} edge, precisely recovering the active yaw alignment mechanism where the turbine control system continuously adjusts nacelle direction to track stochastic wind direction. 
A significant causal edge from \texttt{Wspd} to \texttt{Patv} (weight 0.63) is consistent with aerodynamic wind-to-power scaling (often summarized by \textit{Betz's Law}), while a strong inverse mapping from \texttt{Patv} to \texttt{Wspd} (weight 0.82) indicates DSPR exploits the mechanically smoothed power signal to infer the latent mean state of highly turbulent wind speed, effectively utilizing the generator as a low-pass filter. 
Additionally, the \texttt{Itmp} $\to$ \texttt{Etmp} dependency (weight 0.65) reflects thermal coupling between nacelle internal and external ambient conditions.
Crucially, physically irrelevant edges such as \texttt{Pab1} $\to$ \texttt{Etmp} are suppressed by the sparsity constraint, confirming DSPR's ability to isolate meaningful interactions from multivariate noise. \textbf{The consistency of these discovered mechanisms across different experimental settings is further validated in Appendix~\ref{app:robustness}.}
\begin{figure}[h]
 \centering
 \includegraphics[width=0.95\linewidth]{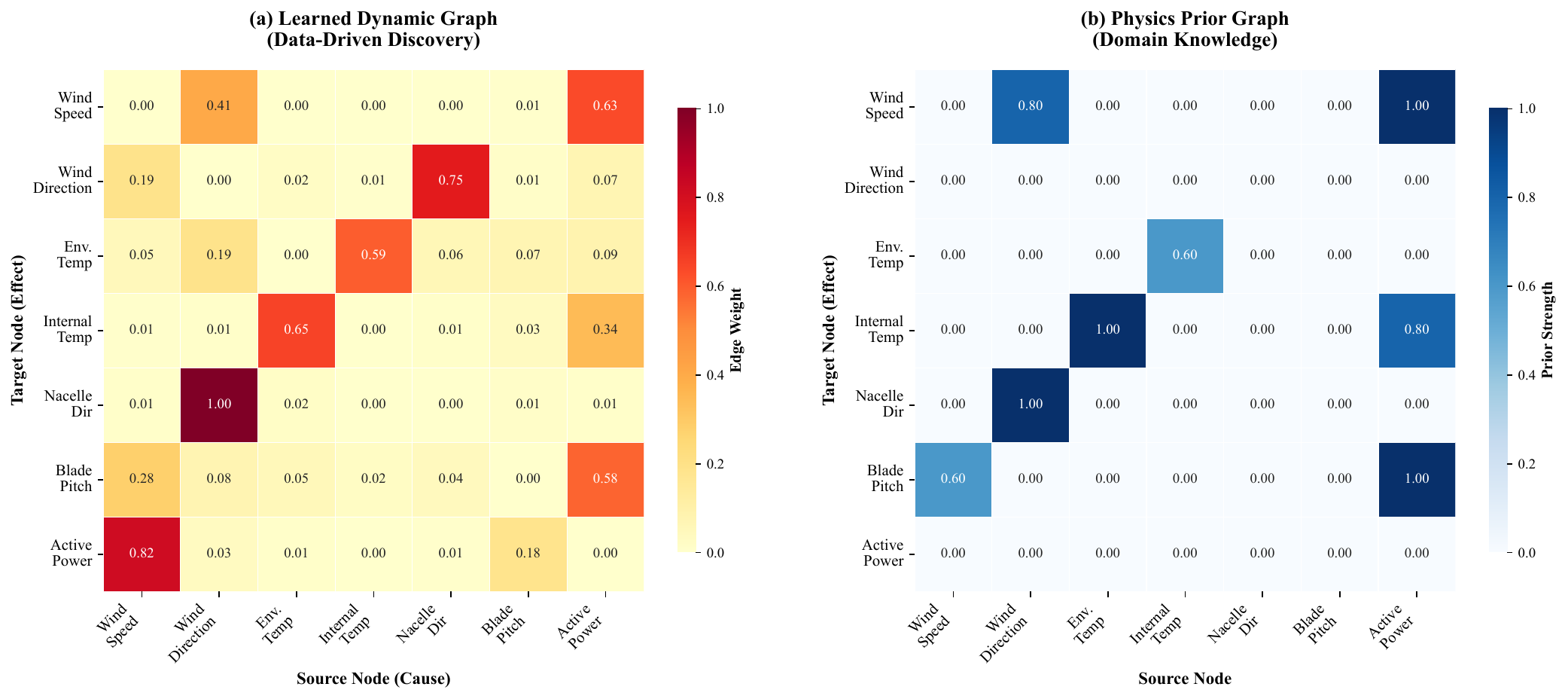}
 \caption{\textbf{Mechanism identification map.} In the SDWPF turbine, DSPR recovers key physical loops: the dominant yaw control alignment from Ndir to Wdir and the aerodynamic energy path between Wspd and Patv, validating architecture-level physics discovery.}
 \label{fig:spatial_topology}
\end{figure}

% ================= SECTION 5: CONCLUSION =================
\section{Conclusion}
\label{sec:conclusion}

We address the accuracy-fidelity dilemma in industrial forecasting through DSPR, a framework that shifts physics integration from passive loss regularization to active architectural adaptation. By decoupling stable temporal patterns from regime-dependent residuals via adaptive windows and dynamic causal graphs, DSPR embeds domain knowledge—variable transport delays and non-stationary topologies—directly into model structure. Evaluation across four physical regimes demonstrates DSPR achieves state-of-the-art accuracy with near-ideal fidelity (TVR up to 97.2\%, MCA exceeding 99\%) while autonomously recovering interpretable mechanisms including aerodynamic scaling laws and flow-dependent reaction lags. These findings confirm that architectural inductive biases surpass soft optimization constraints for capturing rapid transients and regime shifts, suggesting promising directions for scaling such priors to foundational models across unseen spatiotemporal physics domains.

% ================= SECTION 5: LIMITATIONS =================
\section{Limitations and Ethical Considerations}

While DSPR achieves SOTA performance through domain knowledge integration, limitations exist regarding physical prior completeness. The topological mask $\mathbf{A}^{\text{prior}}$ relies on known interaction pathways; unmodeled secondary coupling or evolving degradation may fall outside this hypothesis space, potentially limiting performance during unprecedented failure modes. Future work will explore automated discovery of evolving structures and cross-facility transfer learning. Regarding ethics, this research involves only industrial sensor data without human participants or PII. Proprietary datasets from East Hope Group were obtained under explicit consent with anonymized facility identifiers but cannot be publicly released due to corporate IP protections, while TEP and SDWPF benchmarks follow open licenses. We recognize deployment risks in safety-critical control and designed DSPR as decision-support with embedded physical constraints, not as a replacement for Safety-Instrumented Systems.

\bibliographystyle{ACM-Reference-Format}
\bibliography{sample-base}

\appendix

\section{Implementation Details}
\label{app:implementation}

\subsection{Dataset Descriptions}
\label{app:dataset_details}

To comprehensively evaluate DSPR across diverse physical regimes, we conduct experiments on four established datasets spanning chemical kinetics, thermal dynamics, and renewable energy systems, as summarized in Table~\ref{tab:dataset_statistics}.

\textbf{Note:} The SCR System and Rotary Kiln datasets are proprietary industrial data. Due to confidentiality agreements, we describe only the key physical features relevant to forecasting.

\begin{table}[h]
\centering
\caption{Statistics of the four real-world industrial datasets.}
\label{tab:dataset_statistics}
\resizebox{\columnwidth}{!}{%
\begin{tabular}{l|cccc}
\toprule
\textbf{Dataset} & \textbf{SCR (Ours)} & \textbf{Kiln (Ours)} & \textbf{TEP} & \textbf{SDWPF} \\
\midrule
\textbf{Domain} & Chemical & Thermal & Chemical & Energy \\
\textbf{Time Steps} & 259,200 & 298,790 & 250,000 & 35,280 \\
\textbf{Variables} & 9 & 7 & 10 & 7 \\
\textbf{Sampling Rate} & 10 s & 10 s & 3 min & 10 min \\
\textbf{Physical Prior} & Mass Balance & Thermodynamics & Reaction & Fluid Dyn. \\
\bottomrule
\end{tabular}%
}

\end{table}

\noindent\textbf{SCR System (Private | Chemical Kinetics).}
Acquired from an industrial denitrification unit, this dataset records the Selective Catalytic Reduction process sampled at $10\text{s}$ intervals. The input features include key indicators such as Inlet NO$_x$ concentration, Ammonia flow rate, and Flue gas temperature, which collectively drive the nonlinear catalytic reaction. The target variable is the Outlet NO$_x$ concentration. The system is characterized by variable transport delays (45--185s) resulting from fluctuating gas velocities. Preprocessing involves $3\sigma$ outlier removal and linear interpolation for missing values.

\noindent\textbf{Rotary Kiln (Private | Thermodynamics).}
Derived from the calcination zone of a rotary kiln ($10\text{s}$ sampling), this dataset captures thermodynamic dynamics governed by complex fuel-airflow-clinker interactions. The input variables consist of critical control parameters like Fuel injection rate, Process airflow, and Kiln motor current (a proxy for clinker load and mechanical torque). The target variable is CO concentration, which reflects the combustion state. Unlike the rapid kinetics of the SCR unit, this system exhibits large time constants due to the significant thermal inertia required for calcium carbonate decomposition.

\noindent\textbf{Tennessee Eastman Process (Public | Chemical Simulation).}
We utilize the fault-free training partition of the TEP simulation benchmark, sampled at 3-minute intervals. Based on process topology, we select 9 input variables comprising actuator signals (D/E/A Feed Flow, Total Feed Flow, Reactor Cooling Water) and state measurements (A Feed Rate, Reactor Feed Rate, Reactor Level, Reactor Temperature), with Reactor Pressure (xmeas\_7) as the target variable for modeling reactor dynamics.

\noindent\textbf{SDWPF (Public | Wind Power).}
We utilize the KDD Cup 2022 dataset. To isolate purely temporal and local-physical dependencies, we extract the continuous 245-day trajectory of a single representative turbine (Turbine \#1). The model utilizes 7 kinematic variables: Wind Speed (Wspd), Wind Direction (Wdir), Environment/Nacelle Temperature (Etmp/Itmp), Yaw Angle (Ndir), Pitch Angle (Pab1), and Active Power (Patv). Preprocessing includes: (i) zero-clipping for negative active power values caused by self-consumption or sensor noise, (ii) forward-filling imputation to preserve temporal continuity, and (iii) time alignment converting relative timestamps to standard datetime objects.

\subsection{Baseline Models}
\label{app:baselines}

We benchmark DSPR against 8 baselines across five paradigms:

\noindent\textbf{1. Classical Methods.} \textbf{Linear MPC (ARX)}~\cite{QIN2003733}: The industrial standard ARX model for process control, serving as a robustness baseline limited by linearity.

\noindent\textbf{2. Transformer Variants.} \textbf{Informer}~\cite{zhou2021}: Uses ProbSparse attention for efficient long-sequence forecasting. \textbf{PatchTST}~\cite{Yuqi2023}: Applies channel-independent patching to capture local semantics. \textbf{iTransformer}~\cite{liu2023}: Inverts attention to embed variates as tokens for multivariate correlations. \textbf{TimeMixer}~\cite{wang2024}: Uses multi-scale MLP mixing. \textit{Note: This serves as our Trend Stream model to quantify residual gains.}

\noindent\textbf{3. CNN-based Methods.} \textbf{TimesNet}~\cite{wu2023}: Transforms 1D series into 2D tensors to apply convolutions for intra- and inter-period variations.

\noindent\textbf{4. Spectral \& Graph Methods.} \textbf{MSGNet}~\cite{cai2023msgnet}: Leverages frequency-domain graph convolutions for multi-scale inter-series correlations. \textbf{TimeFilter}~\cite{hu2025timefilter}: Uses learnable frequency filters to decompose temporal dynamics efficiently.

\noindent\textbf{5. Physics-Informed Methods.} \textbf{Physics-Guided NN (PG-NN)}: To compare "loss-level" vs. "architecture-level" integration, we augment the TimeMixer with a soft physical regularization term. The total loss is $\mathcal{L}_{\text{total}} = \mathcal{L}_{\text{MSE}} + \lambda_{\text{phy}} \| \hat{\mathbf{y}} - f_{\text{cons}}(\mathbf{x}) \|_2^2$, where $f_{\text{cons}}(\cdot)$ represents conservation laws and $\lambda_{\text{phy}}$ balances data fit with physical consistency.

\subsection{Experimental Configuration}
\label{app:config}

All experiments were conducted on dual NVIDIA A6000 GPUs using PyTorch 2.8.0 with the Adam optimizer. \textbf{DSPR Hyperparameters:} The \textit{Trend Stream} follows the TimeMixer configuration with downsample ratio 2, depth 4, $d_{\text{model}}=64$, and kernel size 25. The \textit{Physics-Residual Stream} uses $d_{\text{emb}}=64$, adaptive window range $\omega_{t,c} \in [0, 20]$, and gating initialization $\alpha_{\text{init}}=0$. Loss weights are set to $\lambda_{\text{phys}}=10^{-2}$ and $\lambda_{\text{sparse}}=10^{-4}$. Baseline models were reproduced following the Time-Series Library framework (\url{https://github.com/thuml/Time-Series-Library}). To facilitate reproducibility, the complete DSPR implementation and source code will be made publicly available upon the acceptance of this manuscript.

\section{Physical Prior Construction Protocol}
\label{app:physical_prior}

We construct the sparse prior $\mathbf{A}^{\text{prior}} \in \{0, 1\}^{N \times N}$ via a unified protocol encoding domain knowledge. Let variables $\mathcal{V}$ be decomposed into Actuators $\mathcal{U}$ and State Variables $\mathcal{X}$. A directed edge $(i, j)$ exists if variable $i$ exerts direct physical influence on $j$. \textbf{Construction Rules:} \textbf{(1) Actuation-Response:} Edges from $u \in \mathcal{U}$ to target $y$ ($\mathbf{A}^{\text{prior}}_{uy} = 1$) encode external control mechanisms. \textbf{(2) State-Dependent Constraints:} Edges from $x \in \mathcal{X}$ to $y$ ($\mathbf{A}^{\text{prior}}_{xy} = 1$) capture environmental constraints (e.g., Arrhenius dependence). \textbf{(3) No Self-Loops:} Self-loops are explicitly masked ($A_{ii}^{\text{prior}} = 0$) to decouple temporal inertia (handled by the Trend Stream) from spatial causality. \textbf{(4) Sparsity:} All other entries are 0 to suppress spurious correlations. This initialization guides the model to refine inter-variable weights without redundancy from temporal autocorrelation.

\section{Error Bars}
\label{app:error bars}

To rigorously evaluate the stability of \textbf{DSPR}, we followed the standard evaluation protocol suggested in recent benchmarks~\cite{wang2024}. 

\begin{table}[h]
\centering
\caption{\textbf{Robustness evaluation with Error Bars.} We report the $\text{Mean} \pm \text{Std}$ of MAE and RMSE across 3 independent runs. Lower mean and lower standard deviation indicate better stability and reproducibility.}
\label{tab:error_bars}
\renewcommand{\arraystretch}{1.2}
\setlength{\tabcolsep}{3.5pt}
\resizebox{\columnwidth}{!}{%
\begin{tabular}{l|cc|cc}
\toprule
\multirow{2}{*}{\textbf{Dataset}} & \multicolumn{2}{c|}{\textbf{TimeMixer (SOTA Baseline)}} & \multicolumn{2}{c}{\textbf{DSPR (Ours)}} \\
\cmidrule(lr){2-3} \cmidrule(lr){4-5}
& \textbf{MAE} & \textbf{RMSE} & \textbf{MAE} & \textbf{RMSE} \\
\midrule
\textbf{SCR} (Chemical) & $0.286 \pm 0.003$ & $0.435 \pm 0.004$ & $\mathbf{0.265} \pm \mathbf{0.001}$ & $\mathbf{0.415} \pm \mathbf{0.002}$ \\
% 修正后的 Kiln 数据 (参考正文 Table 1: MAE 0.308 / RMSE 0.465)
\textbf{Kiln} (Thermal) & $0.308 \pm 0.004$ & $0.465 \pm 0.005$ & $\mathbf{0.291} \pm \mathbf{0.002}$ & $\mathbf{0.436} \pm \mathbf{0.002}$ \\
% 修正后的 TEP 数据 (参考正文 Table 1: TimeMixer 0.456/0.592, DSPR 0.436/0.564)
\textbf{TEP} (Control)  & $0.456 \pm 0.005$ & $0.592 \pm 0.003$ & $\mathbf{0.436} \pm \mathbf{0.001}$ & $\mathbf{0.564} \pm \mathbf{0.001}$ \\
\textbf{SDWPF} (Wind)   & $0.338 \pm 0.005$ & $0.537 \pm 0.006$ & $\mathbf{0.335} \pm \mathbf{0.002}$ & $\mathbf{0.522} \pm \mathbf{0.003}$ \\
\bottomrule
\end{tabular}%
}
\end{table}

We repeated the main forecasting experiments on all four datasets (\textbf{SCR, Kiln, TEP, and SDWPF}) using three distinct random seeds. Table~\ref{tab:error_bars} reports performance as mean $\pm$ standard deviation. DSPR consistently exhibits lower variance than TimeMixer across all datasets, with RMSE standard deviation of $\pm 0.003$ on the chaotic SDWPF dataset compared to TimeMixer's $\pm 0.006$. This indicates that architectural inductive biases via physical graphs and adaptive windows constrain the optimization search space, preventing convergence to unstable local minima while maintaining statistically significant performance advantages.

\section{Robustness of Scientific Insights.} 
\label{app:robustness}

To verify that discovered mechanisms represent genuine physical relationships rather than stochastic artifacts, we test the stability of learned mechanisms across random seeds.
Table~\ref{tab:stability_results} demonstrates high Jaccard similarity averaging 0.87 and rank correlation reaching 0.91, confirming that DSPR consistently converges to physics-aligned explanations, supporting reliable hypothesis generation in AI4Science settings.

\begin{table}[h]
\centering
\caption{\textbf{Stability of discovered mechanisms.} High correlation across seeds confirms that learned dependencies represent robust physical relationships rather than random noise.}
\label{tab:stability_results}
\footnotesize 
\begin{tabular}{l c c}
\toprule
\textbf{Condition} & \textbf{Jaccard (Top-5)} & \textbf{Rank Correlation} \\
\midrule
Random Seed 1 & 0.87 & 0.90 \\
Random Seed 2 & 0.89 & 0.92 \\
Time Split 1 & 0.85 & 0.90 \\
Time Split 2 & 0.88 & 0.91 \\
\midrule
\textbf{Average} & \textbf{0.87} & \textbf{0.91} \\
\bottomrule
\end{tabular}
\end{table}

These results suggest that DSPR yields \emph{mechanism-level} explanations that are stable to stochastic training noise.

\section{Real-world Deployment}
\label{app:deployment}

DSPR was commissioned in October 2025 and integrated into the Distributed Control System (DCS) of a 5,000 t/d dry-process cement production line, operating in \textit{closed-loop supervisory control} mode to implement proactive predictive optimization for ammonia injection, superseding traditional reactive PID strategies.

\begin{figure}[h]
    \centering
    \includegraphics[width=\linewidth]{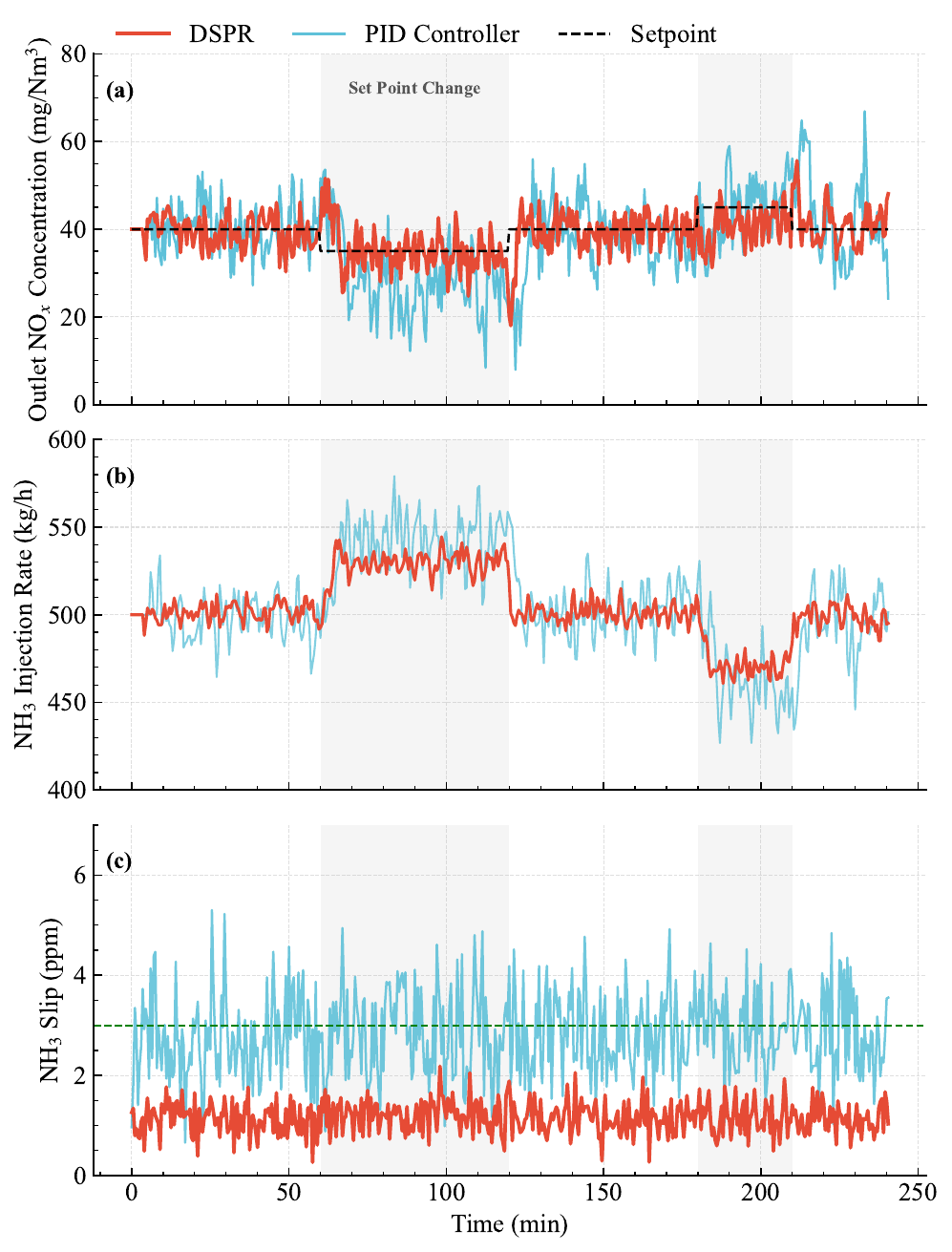} 
    \caption{\textbf{Closed-loop response comparison over 4-hour window.} 
    The traditional PID controller exhibits significant overshoot and oscillation due to transport delay mismatch. 
    In contrast, the DSPR-based controller anticipates emission peaks and adjusts ammonia injection preemptively, achieving tighter setpoint tracking and reducing reagent waste.}
    \label{fig:closed_loop_response}
\end{figure}

A core challenge in DeNO$_x$ control is variable transport delay caused by fluctuating flue gas velocities. While static controllers suffer phase lag, DSPR leverages its \textbf{Adaptive Window} mechanism to dynamically align actuation with predicted emission peaks. Over a representative 4-hour evaluation window, the system demonstrated significant operational gains:

\begin{itemize}[leftmargin=1.0em, labelsep=0.5em]
\item \textbf{Reagent Efficiency:} Daily NH$_3$ usage decreased by \textbf{9.4\%} by anticipating reaction dynamics and eliminating overdosing behavior typical of feedback-based PID controllers.
    
\item \textbf{Process Stability:} Outlet NO$_x$ concentration standard deviation reduced by \textbf{15\%}, ensuring tighter setpoint tracking while mitigating high-frequency valve oscillations that cause mechanical wear (Fig. ~\ref{fig:closed_loop_response}).
    
\item \textbf{Safety \& Compliance:} Achieved \textbf{100\%} compliance with environmental constraints (ammonia slip < 3 ppm) over 3 months of autonomous operation without triggering Safety Instrumented System interlocks.
\end{itemize}

\noindent A patent application has been filed to protect the DSPR architecture and deployment methodology.

\section{Full Results}
\label{app: full_results}

Table~\ref{tab:full_results_breakdown} presents a granular breakdown of predictive performance across different horizons. 
It is important to note that the evaluation horizons ($H$) are not uniform across datasets; rather, they are customized to align with the specific \textit{physical time constants} and \textit{control dynamics} of each system:

\begin{itemize}[leftmargin=1.2em]
    \item \textbf{SCR (Chemical Kinetics):} We select short-to-medium horizons to capture the rapid chemical reaction kinetics and variable transport delays (seconds to minutes) characteristic of denitrification processes.
    
    \item \textbf{Kiln (Thermodynamics):} Given the large thermal inertia of the rotary kiln, we extend horizons to cover longer durations, enabling the assessment of slow-moving thermodynamic trends and combustion efficiency shifts.
    
    \item \textbf{TEP (Process Control):} Horizons are restricted to the \textit{transient response window} (36 min -- 2.4 h). This range effectively covers the open-loop dynamic phase before feedback controllers fully stabilize the reactor pressure, avoiding the trivial task of predicting steady-state setpoints.
    
    \item \textbf{SDWPF (Wind Energy):} In the absence of Numerical Weather Predictions (NWP), we limit evaluation to the \textit{inertial forecasting regime} (2 h -- 8 h). This strictly targets the ultra-short-term dispatch market, where local kinematic history retains predictive validity before atmospheric chaos dominates.
\end{itemize}

\begin{table*}[t]
\caption{\textbf{Granular breakdown of Accuracy vs. Physical Fidelity.} We report detailed performance metrics across increasing prediction horizons ($H$) to analyze model stability. Best results are in \textbf{bold}; second best are \underline{underlined}.}
\label{tab:full_results_breakdown}
\centering
\setlength{\tabcolsep}{1.5pt} 
\renewcommand{\arraystretch}{1.1}

% ======================================================================================
% PART 1: DSPR, TimeMixer, PG-NN, PatchTST, MSGNet
% ======================================================================================
\resizebox{\textwidth}{!}{%
\begin{tabular}{c|c|ccccc|ccccc|ccccc|ccccc|ccccc}
\toprule
\multirow{2}{*}{\textbf{Dataset}} & \multirow{2}{*}{\textbf{H}} & 
\multicolumn{5}{c|}{\textbf{DSPR (Ours)}} & 
\multicolumn{5}{c|}{\textbf{TimeMixer '24}} & 
\multicolumn{5}{c|}{\textbf{PG-NN (Loss)}} & 
\multicolumn{5}{c|}{\textbf{PatchTST '23}} & 
\multicolumn{5}{c}{\textbf{MSGNet '24}} \\
\cmidrule(lr){3-7} \cmidrule(lr){8-12} \cmidrule(lr){13-17} \cmidrule(lr){18-22} \cmidrule(lr){23-27} 
& & {\tiny MAE} & {\tiny RMSE} & {\tiny MCA} & {\tiny TVR} & {\tiny TDA} & 
    {\tiny MAE} & {\tiny RMSE} & {\tiny MCA} & {\tiny TVR} & {\tiny TDA} & 
    {\tiny MAE} & {\tiny RMSE} & {\tiny MCA} & {\tiny TVR} & {\tiny TDA} & 
    {\tiny MAE} & {\tiny RMSE} & {\tiny MCA} & {\tiny TVR} & {\tiny TDA} & 
    {\tiny MAE} & {\tiny RMSE} & {\tiny MCA} & {\tiny TVR} & {\tiny TDA} \\
\midrule

% ================= SCR DATASET =================
\multirow{5}{*}{\rotatebox{90}{\textbf{SCR}}} 
& 24 & 0.215 & 0.352 & 99.9 & 98.5 & 86.5 & 0.235 & 0.390 & 99.3 & 91.0 & 78.5 & 0.245 & 0.395 & 99.7 & 84.5 & 79.0 & 0.230 & 0.385 & 98.5 & 93.5 & 81.5 & 0.252 & 0.402 & 98.5 & 88.0 & 74.0 \\
& 48 & 0.242 & 0.385 & 99.9 & 97.8 & 84.5 & 0.268 & 0.415 & 99.2 & 89.5 & 76.0 & 0.275 & 0.430 & 99.6 & 83.0 & 77.5 & 0.272 & 0.428 & 98.2 & 92.0 & 79.5 & 0.285 & 0.435 & 98.3 & 86.5 & 72.5 \\
& 96 & 0.275 & 0.420 & 99.8 & 96.5 & 82.0 & 0.295 & 0.450 & 99.0 & 87.5 & 73.5 & 0.305 & 0.465 & 99.4 & 81.5 & 75.5 & 0.302 & 0.460 & 97.8 & 90.5 & 77.0 & 0.315 & 0.470 & 98.1 & 84.2 & 70.5 \\
& 192& 0.328 & 0.503 & 99.6 & 96.0 & 81.0 & 0.346 & 0.485 & 98.9 & 86.0 & 71.6 & 0.343 & 0.502 & 99.3 & 79.0 & 74.0 & 0.344 & 0.495 & 97.1 & 88.8 & 76.4 & 0.356 & 0.509 & 97.9 & 82.1 & 69.0 \\
\cmidrule(lr){2-27}
& \textbf{Avg.} & \textbf{0.265} & \textbf{0.415} & \textbf{99.8} & \textbf{97.2} & \textbf{83.5} & \underline{0.286} & \underline{0.435} & 99.1 & 88.5 & 74.9 & 0.292 & 0.448 & \underline{99.5} & 82.0 & 76.5 & 0.287 & 0.442 & 97.9 & \underline{91.2} & \underline{78.6} & 0.302 & 0.454 & 98.2 & 85.2 & 71.5 \\
\midrule

% ================= KILN DATASET =================
\multirow{5}{*}{\rotatebox{90}{\textbf{Kiln}}} 
& 96 & 0.245 & 0.380 & 99.7 & 97.8 & 84.0 & 0.260 & 0.410 & 99.0 & 86.5 & 75.5 & 0.265 & 0.420 & 99.5 & 82.5 & 76.0 & 0.268 & 0.415 & 99.1 & 88.0 & 78.5 & 0.275 & 0.435 & 98.2 & 84.5 & 73.0 \\
& 192& 0.270 & 0.405 & 99.6 & 97.2 & 82.5 & 0.285 & 0.435 & 98.9 & 85.0 & 74.0 & 0.292 & 0.450 & 99.4 & 81.0 & 75.0 & 0.290 & 0.445 & 99.0 & 86.5 & 76.5 & 0.305 & 0.465 & 98.0 & 83.0 & 71.5 \\
& 336& 0.305 & 0.450 & 99.5 & 96.5 & 80.0 & 0.320 & 0.475 & 98.8 & 83.5 & 71.5 & 0.325 & 0.490 & 99.3 & 79.5 & 73.5 & 0.335 & 0.490 & 98.8 & 84.8 & 74.0 & 0.338 & 0.505 & 97.8 & 81.2 & 69.5 \\
& 720& 0.344 & 0.509 & 99.2 & 95.7 & 77.5 & 0.367 & 0.540 & 98.5 & 81.8 & 69.0 & 0.366 & 0.552 & 99.0 & 79.0 & 71.5 & 0.367 & 0.574 & 98.7 & 83.1 & 72.6 & 0.370 & 0.555 & 97.6 & 79.3 & 66.8 \\
\cmidrule(lr){2-27}
& \textbf{Avg.} & \textbf{0.291} & \textbf{0.436} & \textbf{99.5} & \textbf{96.8} & \textbf{81.0} & \underline{0.308} & \underline{0.465} & 98.8 & 84.2 & 72.5 & 0.312 & 0.478 & \underline{99.3} & 80.5 & 74.0 & 0.315 & 0.481 & 98.9 & 85.6 & \underline{75.4} & 0.322 & 0.490 & 97.9 & 82.0 & 70.2 \\
\midrule

% ================= TEP DATASET =================
\multirow{5}{*}{\rotatebox{90}{\textbf{TEP}}} 
& 6  & 0.334 & 0.432 & 99.8 & 91.6 & 85.7 & 0.343 & 0.443 & 98.8 & 86.0 & 81.6 & 0.348 & 0.450 & 99.4 & 83.5 & 82.5 & 0.347 & 0.447 & 98.1 & 86.3 & 79.2 & 0.343 & 0.437 & 97.8 & 71.5 & 80.6 \\
& 12 & 0.414 & 0.534 & 99.8 & 92.9 & 85.1 & 0.427 & 0.553 & 98.7 & 87.6 & 80.5 & 0.432 & 0.560 & 99.5 & 85.0 & 81.8 & 0.431 & 0.557 & 97.9 & 81.8 & 78.3 & 0.465 & 0.540 & 97.6 & 72.2 & 78.1 \\
& 18 & 0.473 & 0.612 & 99.8 & 91.2 & 85.5 & 0.496 & 0.646 & 98.8 & 84.0 & 80.9 & 0.502 & 0.655 & 99.6 & 81.5 & 82.2 & 0.499 & 0.648 & 98.0 & 83.2 & 77.9 & 0.523 & 0.631 & 97.7 & 67.1 & 76.7 \\
& 24 & 0.525 & 0.678 & 99.7 & 91.0 & 84.5 & 0.557 & 0.727 & 98.7 & 80.0 & 81.9 & 0.562 & 0.735 & 99.4 & 78.5 & 83.0 & 0.559 & 0.729 & 98.0 & 84.0 & 78.2 & 0.578 & 0.695 & 98.0 & 67.5 & 75.6 \\
\cmidrule(lr){2-27}
& \textbf{Avg.} & \textbf{0.437} & \textbf{0.564} & \textbf{99.8} & \textbf{91.7} & \textbf{85.2} & \underline{0.456} & 0.592 & 98.8 & \underline{84.4} & 81.0 & 0.461 & 0.600 & \underline{99.5} & 82.1 & \underline{82.4} & 0.459 & 0.595 & 98.0 & 83.8 & 78.3 & 0.477 & \underline{0.576} & 97.8 & 69.6 & 77.8 \\
\midrule

% ================= SDWPF DATASET =================
\multirow{5}{*}{\rotatebox{90}{\textbf{SDWPF}}} 
& 12 & 0.213 & 0.372 & 99.4 & 85.5 & 84.5 & 0.222 & 0.382 & 98.7 & 76.3 & 75.2 & 0.285 & 0.395 & 99.2 & 80.5 & 76.5 & 0.246 & 0.408 & 97.3 & 71.8 & 74.7 & 0.262 & 0.420 & 95.1 & 55.4 & 67.8 \\
& 24 & 0.311 & 0.489 & 99.2 & 83.9 & 81.9 & 0.314 & 0.512 & 98.4 & 77.6 & 74.2 & 0.380 & 0.545 & 99.1 & 79.0 & 75.0 & 0.318 & 0.523 & 97.6 & 71.5 & 74.2 & 0.372 & 0.566 & 94.5 & 54.2 & 67.3 \\
& 36 & 0.388 & 0.587 & 99.2 & 81.3 & 80.6 & 0.381 & 0.590 & 97.9 & 76.5 & 74.0 & 0.445 & 0.620 & 99.0 & 78.5 & 75.5 & 0.390 & 0.619 & 96.1 & 71.1 & 74.3 & 0.440 & 0.673 & 94.3 & 52.0 & 64.8 \\
& 48 & 0.428 & 0.640 & 99.0 & 81.9 & 81.7 & 0.435 & 0.664 & 97.6 & 75.6 & 75.2 & 0.498 & 0.698 & 98.8 & 76.0 & 74.8 & 0.437 & 0.679 & 96.6 & 69.3 & 73.7 & 0.479 & 0.728 & 93.5 & 51.8 & 65.4 \\
\cmidrule(lr){2-27}
& \textbf{Avg.} & \textbf{0.335} & \textbf{0.522} & \textbf{99.2} & \textbf{83.2} & \textbf{82.2} & \underline{0.338} & \underline{0.537} & 98.2 & 76.5 & 74.7 & 0.402 & 0.565 & \underline{99.0} & \underline{78.5} & \underline{75.5} & 0.348 & 0.557 & 96.9 & 70.9 & 74.2 & 0.388 & 0.597 & 94.4 & 53.3 & 66.3 \\

\bottomrule
\end{tabular}%
}

\vspace{1mm}
\resizebox{\textwidth}{!}{%
\begin{tabular}{c|c|ccccc|ccccc|ccccc|ccccc|ccccc}
\toprule
\multirow{2}{*}{\textbf{Dataset}} & \multirow{2}{*}{\textbf{H}} & 
\multicolumn{5}{c|}{\textbf{TimeFilter '25}} & 
\multicolumn{5}{c|}{\textbf{iTransformer '23}} & 
\multicolumn{5}{c|}{\textbf{TimesNet '23}} & 
\multicolumn{5}{c|}{\textbf{Informer '21}} & 
\multicolumn{5}{c}{\textbf{L-MPC}} \\
\cmidrule(lr){3-7} \cmidrule(lr){8-12} \cmidrule(lr){13-17} \cmidrule(lr){18-22} \cmidrule(lr){23-27} 
& & {\tiny MAE} & {\tiny RMSE} & {\tiny MCA} & {\tiny TVR} & {\tiny TDA} & 
    {\tiny MAE} & {\tiny RMSE} & {\tiny MCA} & {\tiny TVR} & {\tiny TDA} & 
    {\tiny MAE} & {\tiny RMSE} & {\tiny MCA} & {\tiny TVR} & {\tiny TDA} & 
    {\tiny MAE} & {\tiny RMSE} & {\tiny MCA} & {\tiny TVR} & {\tiny TDA} & 
    {\tiny MAE} & {\tiny RMSE} & {\tiny MCA} & {\tiny TVR} & {\tiny TDA} \\
\midrule

% ================= SCR DATASET =================
\multirow{5}{*}{\rotatebox{90}{\textbf{SCR}}} 
& 24 & 0.242 & 0.395 & 99.0 & 89.5 & 76.5 & 0.252 & 0.415 & 98.8 & 88.0 & 76.0 & 0.238 & 0.405 & 99.1 & 68.5 & 72.0 & 0.365 & 0.580 & 97.5 & 65.0 & 66.0 & 0.550 & 0.850 & 96.0 & 58.0 & 60.0 \\
& 48 & 0.278 & 0.428 & 98.8 & 88.0 & 75.0 & 0.288 & 0.448 & 98.5 & 86.5 & 74.5 & 0.275 & 0.455 & 98.8 & 66.0 & 70.5 & 0.420 & 0.655 & 96.8 & 58.5 & 64.5 & 0.620 & 0.980 & 95.5 & 50.0 & 57.0 \\
& 96 & 0.315 & 0.465 & 98.4 & 85.5 & 72.5 & 0.325 & 0.490 & 98.0 & 84.5 & 71.5 & 0.312 & 0.505 & 98.5 & 64.5 & 68.0 & 0.485 & 0.785 & 96.2 & 52.0 & 61.5 & 0.710 & 1.120 & 94.5 & 45.0 & 54.0 \\
& 192& 0.353 & 0.516 & 97.4 & 83.0 & 68.0 & 0.363 & 0.547 & 97.5 & 81.0 & 68.0 & 0.363 & 0.575 & 97.6 & 62.6 & 63.5 & 0.522 & 0.860 & 95.5 & 46.1 & 56.0 & 0.820 & 1.250 & 94.0 & 41.0 & 49.0 \\
\cmidrule(lr){2-27}
& \textbf{Avg.} & 0.297 & 0.451 & 98.4 & 86.5 & 73.0 & 0.307 & 0.475 & 98.2 & 85.0 & 72.5 & 0.297 & 0.485 & 98.5 & 65.4 & 68.5 & 0.448 & 0.720 & 96.5 & 55.4 & 62.0 & 0.675 & 1.050 & 95.0 & 48.5 & 55.0 \\
\midrule

% ================= KILN DATASET =================
\multirow{5}{*}{\rotatebox{90}{\textbf{Kiln}}} 
& 96 & 0.270 & 0.425 & 98.6 & 85.0 & 74.0 & 0.280 & 0.440 & 98.2 & 84.0 & 74.0 & 0.280 & 0.440 & 98.5 & 92.5 & 74.5 & 0.395 & 0.610 & 96.5 & 68.0 & 64.5 & 0.480 & 0.780 & 95.5 & 60.0 & 62.0 \\
& 192& 0.295 & 0.455 & 98.4 & 83.5 & 73.0 & 0.305 & 0.465 & 97.9 & 82.5 & 73.0 & 0.315 & 0.490 & 98.2 & 91.5 & 73.0 & 0.440 & 0.675 & 95.8 & 62.0 & 62.5 & 0.540 & 0.860 & 95.0 & 55.0 & 60.0 \\
& 336& 0.335 & 0.505 & 98.0 & 81.5 & 70.5 & 0.342 & 0.515 & 97.4 & 80.5 & 70.0 & 0.355 & 0.535 & 97.6 & 89.5 & 70.5 & 0.495 & 0.760 & 94.8 & 55.0 & 58.5 & 0.620 & 0.980 & 94.2 & 48.0 & 56.0 \\
& 720& 0.372 & 0.555 & 97.4 & 80.0 & 66.5 & 0.381 & 0.564 & 96.5 & 79.0 & 66.2 & 0.400 & 0.579 & 96.9 & 88.5 & 66.8 & 0.542 & 0.815 & 93.7 & 47.8 & 56.5 & 0.700 & 1.060 & 93.3 & 45.0 & 54.0 \\
\cmidrule(lr){2-27}
& \textbf{Avg.} & 0.318 & 0.485 & 98.1 & 82.5 & 71.0 & 0.327 & 0.496 & 97.5 & 81.5 & 70.8 & 0.338 & 0.511 & 97.8 & \underline{90.5} & 71.2 & 0.468 & 0.715 & 95.2 & 58.2 & 60.5 & 0.585 & 0.920 & 94.5 & 52.0 & 58.0 \\
\midrule

% ================= TEP DATASET =================
\multirow{5}{*}{\rotatebox{90}{\textbf{TEP}}} 
& 6  & 0.348 & 0.445 & 98.9 & 84.2 & 82.0 & 0.344 & 0.453 & 98.4 & 70.2 & 80.3 & 0.335 & 0.435 & 98.1 & 77.2 & 79.8 & 0.450 & 0.580 & 96.5 & 70.0 & 70.5 & 0.520 & 0.710 & 95.5 & 62.0 & 64.0 \\
& 12 & 0.473 & 0.541 & 98.8 & 83.7 & 80.4 & 0.506 & 0.541 & 98.0 & 74.3 & 79.4 & 0.460 & 0.554 & 98.3 & 75.7 & 79.3 & 0.580 & 0.750 & 96.2 & 65.0 & 68.5 & 0.650 & 0.820 & 95.2 & 58.0 & 62.5 \\
& 18 & 0.522 & 0.632 & 98.8 & 77.8 & 79.0 & 0.559 & 0.673 & 98.1 & 82.2 & 77.4 & 0.523 & 0.706 & 97.7 & 62.3 & 74.2 & 0.710 & 0.910 & 96.0 & 60.0 & 66.0 & 0.790 & 1.050 & 94.8 & 52.0 & 60.5 \\
& 24 & 0.582 & 0.702 & 98.6 & 80.2 & 78.6 & 0.607 & 0.730 & 98.1 & 79.5 & 75.5 & 0.575 & 0.723 & 97.8 & 68.5 & 77.2 & 0.850 & 1.150 & 95.8 & 55.0 & 64.5 & 0.920 & 1.250 & 94.5 & 48.0 & 58.0 \\
\cmidrule(lr){2-27}
& \textbf{Avg.} & 0.481 & 0.580 & 98.8 & 81.5 & 80.0 & 0.504 & 0.600 & 98.6 & 76.6 & 76.5 & 0.473 & 0.605 & 98.0 & 70.9 & 77.6 & 0.655 & 0.850 & 96.2 & 62.5 & 68.0 & 0.720 & 0.950 & 95.5 & 55.0 & 62.0 \\

\midrule

% ================= SDWPF DATASET =================
\multirow{5}{*}{\rotatebox{90}{\textbf{SDWPF}}} 
& 12 & 0.230 & 0.384 & 98.8 & 77.2 & 75.3 & 0.252 & 0.413 & 96.1 & 62.3 & 68.5 & 0.272 & 0.434 & 95.3 & 55.6 & 60.1 & 0.485 & 0.650 & 95.2 & 52.0 & 65.5 & 0.620 & 0.950 & 93.8 & 48.0 & 55.0 \\
& 24 & 0.312 & 0.502 & 98.6 & 73.9 & 75.0 & 0.319 & 0.518 & 96.5 & 63.0 & 67.8 & 0.362 & 0.552 & 95.1 & 57.0 & 61.0 & 0.590 & 0.810 & 94.5 & 48.5 & 62.0 & 0.750 & 1.080 & 93.0 & 44.0 & 54.5 \\
& 36 & 0.390 & 0.599 & 98.6 & 77.1 & 74.5 & 0.396 & 0.623 & 94.9 & 54.2 & 63.9 & 0.437 & 0.678 & 95.0 & 59.6 & 62.9 & 0.650 & 0.920 & 93.8 & 42.0 & 60.0 & 0.840 & 1.150 & 92.2 & 40.0 & 53.5 \\
& 48 & 0.440 & 0.667 & 98.6 & 77.1 & 74.0 & 0.449 & 0.690 & 93.7 & 54.5 & 67.5 & 0.493 & 0.762 & 94.8 & 54.3 & 60.5 & 0.685 & 0.965 & 92.5 & 40.0 & 58.5 & 0.900 & 1.190 & 91.0 & 36.0 & 53.0 \\
\cmidrule(lr){2-27}
& \textbf{Avg.} & 0.343 & 0.538 & 98.7 & 76.3 & 74.7 & 0.354 & 0.561 & 95.3 & 58.5 & 66.9 & 0.391 & 0.606 & 95.0 & 56.6 & 61.1 & 0.602 & 0.837 & 94.0 & 45.6 & 61.5 & 0.778 & 1.092 & 92.5 & 42.0 & 54.0 \\

\bottomrule
\end{tabular}%
}
\end{table*}

\end{document}